\documentclass[journal]{IEEEtran}
\usepackage{xcolor}
\usepackage{subcaption}
\usepackage{graphicx}
\usepackage{url}
\usepackage{bm}
\usepackage{amssymb}
\usepackage{amsmath}
\usepackage{algorithm}
\usepackage{algorithmic}
\usepackage{multirow}
\usepackage{array}
\usepackage[most]{tcolorbox}
\usepackage{lipsum}
\usepackage{listings}
\usepackage[hidelinks]{hyperref}
\usepackage{pifont}
\usepackage{makecell}

\definecolor{codegreen}{rgb}{0,0.6,0}
\definecolor{codegray}{rgb}{0.5,0.5,0.5}
\definecolor{codepurple}{rgb}{0.58,0,0.82}
\definecolor{backcolour}{rgb}{0.95,0.95,0.95}

\lstdefinestyle{mystyle}{
  backgroundcolor=\color{backcolour},
  commentstyle=\color{codegreen},
  keywordstyle=\color{magenta},
  numberstyle=\tiny\color{codegray},
  stringstyle=\color{codepurple},
  basicstyle=\ttfamily\footnotesize,
  breakatwhitespace=false,         
  breaklines=true,                 
  captionpos=b,                    
  keepspaces=true,                 
  numbers=left,                    
  numbersep=5pt,                  
  showspaces=false,                
  showstringspaces=false,
  showtabs=false,                  
  tabsize=2
}

\usepackage{graphicx}

\newcommand*{\img}[1]{%
    \raisebox{-.3\baselineskip}{%
        \includegraphics[
        height=\baselineskip,
        width=\baselineskip,
        keepaspectratio,
        ]{#1}%
    }%
}

\newcommand{\figuretag}[1]{%
  \addtocounter{figure}{-1}%
  \renewcommand{\thefigure}{#1}%
}

\newcommand{\tabletag}[1]{%
  \addtocounter{table}{-1}%
  \renewcommand{\thetable}{#1}%
}

\newcommand{\algorithmtag}[1]{%
  \addtocounter{algorithm}{-1}%
  \renewcommand{\thealgorithm}{#1}%
}

\usepackage[normalem]{ulem}
\usepackage{enumitem}

\ifCLASSINFOpdf
\else
\fi
\begin{document}
\setlength{\arraycolsep}{1.5pt}
\title{RLingua: Improving Reinforcement Learning Sample Efficiency in Robotic Manipulations \\With Large Language Models}

\author{Liangliang Chen, Yutian Lei, Shiyu Jin, Ying Zhang, \IEEEmembership{Senior Member, IEEE,} and Liangjun Zhang
\thanks{Liangliang Chen is with the Robotics and Autonomous Driving Lab, Baidu Research, Sunnyvale, CA 94089 USA, and also with the School of Electrical and Computer Engineering, Georgia Institute of Technology, Atlanta,
		GA 30332 USA (e-mail: liangliang.chen@gatech.edu).}
\thanks{Yutian Lei, Shiyu Jin, and Liangjun Zhang are with the Robotics and Autonomous Driving Lab, Baidu Research, Sunnyvale, CA 94089 USA (email: yutianlei@baidu.com, shiyujin@baidu.com, liangjunzhang@baidu.com).}
\thanks{Ying Zhang is with the School of Electrical and Computer Engineering, Georgia Institute of Technology, Atlanta,
		GA 30332 USA (e-mail: yzhang@gatech.edu).}
}

\maketitle
\begin{abstract}
Reinforcement learning (RL) has demonstrated its capability in solving various tasks but is notorious for its low sample efficiency. In this paper, we propose RLingua, a framework that can leverage the internal knowledge of large language models (LLMs) to reduce the sample complexity of RL in robotic manipulations. {\color{black}To this end, we first present a method for extracting the prior knowledge of LLMs by prompt engineering so that a preliminary rule-based robot controller for a specific task can be generated in a user-friendly manner.} Despite being imperfect, the LLM-generated robot controller is utilized to produce action samples during rollouts with a decaying probability, thereby improving RL's sample efficiency. We employ TD3, the widely-used RL baseline method, and modify the actor loss to regularize the policy learning towards the LLM-generated controller. RLingua also provides a novel method of improving the imperfect LLM-generated robot controllers by RL. {\color{black}We demonstrate that RLingua can significantly reduce the sample complexity of TD3 in four robot tasks of {\texttt{panda\_gym}} and achieve high success rates in 12 sampled sparsely rewarded robot tasks in {\texttt{RLBench}}, where the standard TD3 fails.} Additionally, We validated RLingua's effectiveness in real-world robot experiments through Sim2Real, demonstrating that the learned policies are effectively transferable to real robot tasks. Further details about our work are available at our project website {\url{https://rlingua.github.io}}.
\end{abstract}


\IEEEpeerreviewmaketitle

\begin{figure*}[!t]
\centering
\includegraphics[width=0.89\textwidth]{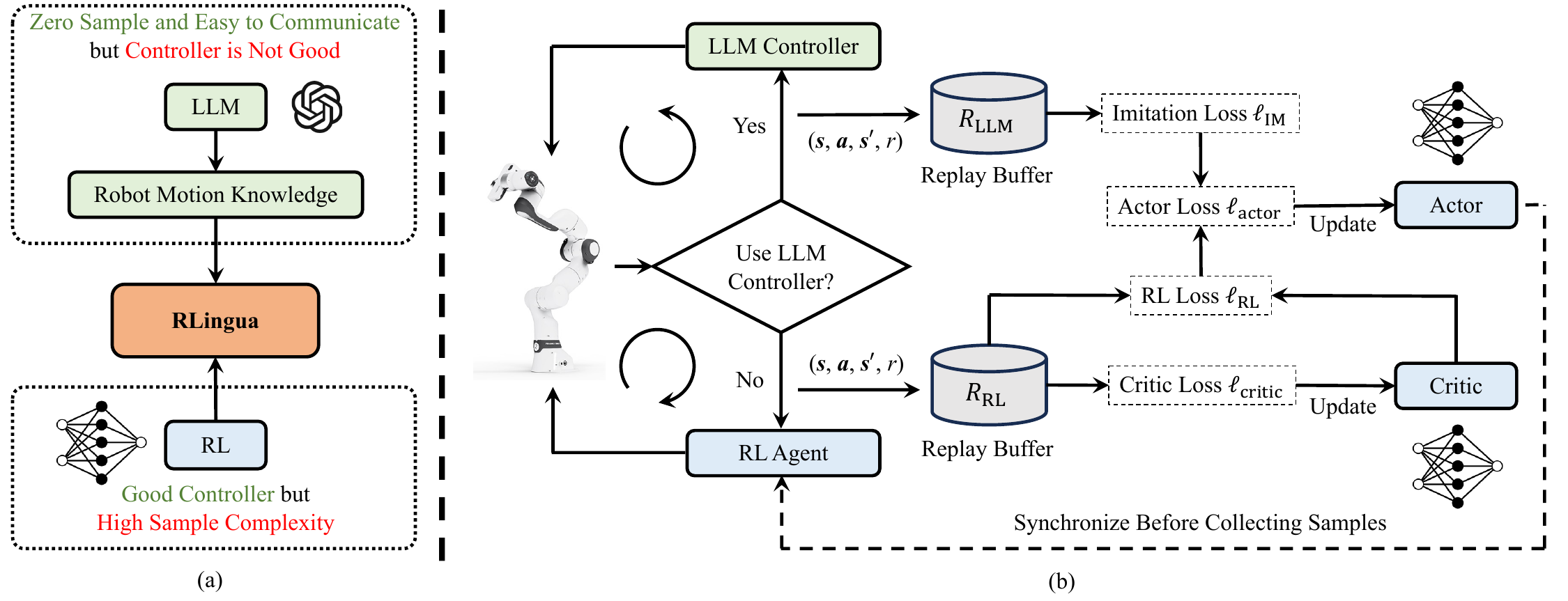}
\caption{RLingua extracts the LLM's knowledge about robot motion to improve the sample efficiency of RL. (a) Motivation: LLMs do not need environment samples and are easy to communicate for non-experts. However, the robot controllers generated directly by LLMs may have inferior performance. In contrast, RL can be used to train robot controllers to achieve high performance. However, the cost of RL is its high sample complexity. (b) Framework: RLingua extracts the internal knowledge of LLMs about robot motion to a coded imperfect controller, which is then used to collect data by interaction with the environment. The robot control policy is trained with both the collected LLM demonstration data and the interaction data collected by the online training policy. {\color{black}The collected demonstration data end up in the LLM buffer $R_{\mathrm{LLM}}$ and are {\textit{directly}} used in imitation learning. They influence RL {\textit{indirectly}} by shaping the policy and thus changing the collected samples in the RL buffer $R_{\mathrm{RL}}$.}}
\label{Fig5}
\end{figure*}

\section{Introduction}
\IEEEPARstart{R}{einforcement} learning (RL) has been demonstrated being able to effectively solve various tasks, ranging from Atari video games \cite{mnih2015human}, chess, shogi, and Go \cite{silver2018general}, to robotic manipulations \cite{lillicrap2015continuous, fan2018surreal}. However, there are some intrinsic issues associated with RL, including the difficulty of reward engineering and low data efficiency in real-world applications \cite{dulac2019challenges}. 

The rapid development of large language models (LLMs) has enabled us to obtain human-level responses across a broad range of professional fields, including history, math, engineering, etc., by providing appropriate prompts \cite{ouyang2022training, openai2023gpt4}. In the context of RL, some researchers have explored the LLMs' potential to generate reward functions \cite{kwon2023reward, yu2023language, xie2023text2reward, ma2023eureka}. The basic idea of these studies is to leverage the prior knowledge of LLMs to generate reward functions for RL algorithms. To this end, the authors use the in-context learning capacities of LLMs or incorporate human interactions into reward generation. Once generated, the reward functions are used by standard RL or planning algorithms to develop control policies in various scenarios. However, these algorithms often require a large number of interaction samples to achieve high performance \cite{xie2023text2reward}. {\textit{In this paper, we aim to explore how LLMs can help reduce the sample complexity of RL algorithms.}} We note that our approach is complementary to the reward design methods in \cite{kwon2023reward, yu2023language, xie2023text2reward, ma2023eureka}. Future algorithms could potentially integrate both approaches to maximize benefits. 

One important cause of the low sample efficiency of RL algorithms is that the agents learn from scratch by trial-and-error \cite{sutton2018reinforcement}. During the learning process, the agent explores the environment gradually, in which many samples need to be collected from the environment. This issue is amplified in environments with sparse rewards, where agents must take more exploration steps to receive non-zero reward signals. We can seek to reduce the sample complexity by guiding the exploration process so that the policy can quickly find a mediocre policy. Then, the policy can continue collecting more data with non-zero rewards and the policy training process can be boosted further. In this context, researchers proposed deep RL from demonstrations \cite{hester2018deep, vecerik2017leveraging, nair2018overcoming}. 
The authors in \cite{hester2018deep, vecerik2017leveraging} combined the DQN \cite{mnih2015human} and DDPG \cite{lillicrap2015continuous} algorithms with learning from demonstrations. Before training begins, the replay buffers in these studies are preloaded with demonstration data, ideally from expert sources. Nair {\textit{et al.}} \cite{nair2018overcoming} leveraged two replay buffers to separately store the static demonstration data and interaction data collected during policy updating. The optimized loss function is a linear combination of behavior cloning and RL losses. However, a common limitation of these methods is their dependence on pre-collected demonstration datasets, typically gathered by human experts. Note that the expert data collection is often labor-intensive and not always feasible. This naturally leads to a question: {\textit{Is it possible to bypass the need for expert-collected demonstration data by utilizing LLMs to generate the data instead?}}

Considering LLMs as repositories of extensive general knowledge \cite{wei2022chain, zhu2023large, mirchandani2023large}, we can utilize them to reduce the need for extensive data collection. Some researchers have leveraged LLMs to assist policy learning in various tasks. The authors in \cite{huang2022language} employed LLMs to generate high-level motions for various tasks in a zero-shot manner. They demonstrated that LLMs can logically break down a multi-stage task into actionable sub-tasks. Ahn {\textit{et al.}} \cite{ahn2022can} expanded this by further considering robots' capabilities under the current stage and learning a value function for action primitives. The study in \cite{huang2022inner} enhanced LLM planners with additional feedback sources, such as success detection, scene detector, and human interaction, enriching robotic control scenarios for LLM planners. Zhang {\textit{et al.}} \cite{zhang2023bootstrap} showed how chaining action primitives can form new, longer-horizon tasks with the assistance of LLM planning.  
However, these methods all assumed the availability of pre-trained action primitives and did not address how to obtain these from lower-level actions. {\color{black}Liang {\textit{et al.}} \cite{liang2023code} directly generated Python codes for robot controllers in terms of both low-level primitives and high-level planners. The authors demonstrated that their framework, ``code as policies", can generate both reactive and way-point policies. Huang {\textit{et al.}} \cite{huang2023voxposer} exploited the code-writing and open-world reasoning abilities of LLMs and generalizable visual grounding capacities of vision-language models to compose 3D value maps in observation space of the agent for guiding robotic interactions.} To the best of our knowledge, there is no prior work focusing on building an RL policy from low-level basic actions with the assistance of LLMs. 

In this paper, we propose RLingua (\textbf{R}einforcement \textbf{L}earn-\textbf{ing} with the guidance of Lang{\textbf{ua}}ge models), a novel framework that trains RL agents with the guidance of language models. RLingua harnesses the extensive prior knowledge embedded in LLMs about robot motions and coding to significantly enhance RL processes. This is achieved by reducing RL sample complexities and refining imperfect robot controllers generated by LLMs through RL. The main contributions are listed as follows.

\begin{itemize}
\item [i)] We introduce a method to generate rule-based controllers through LLM prompts, enabling robots to perform high-level tasks despite only executing low-level movements. This approach differs from existing methods as it does not rely on high-level primitives. {\color{black} In addition, the prompts are designed so that non-experts can prompt the method with natural language instructions.}
\item [ii)] RLingua uses controllers generated by LLMs to collect training data through environment interaction. This novel approach significantly reduces the number of samples needed for effective RL. 
\item [iii)] We validate RLingua's effectiveness in various robotic manipulation tasks with sparse rewards. Our results show a notable reduction in sample complexity and high success rates in tasks challenging for standard RL methods. Furthermore, we demonstrate the practical deployment of the learned RL policy on real robots.
\end{itemize}

\section{Preliminaries}
\subsection{Reinforcement Learning}
Let us consider a Markov decision process (MDP) described by $\mathcal{M}\triangleq(\mathcal{S},\mathcal{A},p_{0},p,r,\gamma)$ \cite{sutton2018reinforcement}. In this tuple $\mathcal{M}$, $\mathcal{S}$ and $\mathcal{A}$ represent the states of state $\bm{s}$ and action $\bm{a}$, respectively, $p_{0}(\bm{s}):\mathcal{S}\mapsto[0, 1]$ denotes the distribution of initial state, $p(\bm{s}^{\prime}\mid\bm{s},\bm{a}):\mathcal{S}\times\mathcal{A}\mapsto{\color{black}\Delta(\mathcal{S})}$ denotes the state transition probability with $\Delta(\cdot)$ being the simplex notation, $r(\bm{s},\bm{a},\bm{s}^{\prime}):\mathcal{S}\times\mathcal{A}\times\mathcal{S}\mapsto\{0, 1\}$ is a sparse reward function indicating if the task is completed or not, and $\gamma\in[0,1]$ denotes the reward discount factor. The goal of RL is to maximize
\begin{eqnarray}
J(\pi\mid\mathcal{M})\triangleq \sum_{t=0}^{\infty}&&\mathbb{E}_{\bm{s}_{0}\sim p_{0},\bm{s}_{t+1}\sim p(\cdot\mid\bm{s}_{t},\bm{a}_{t}),\bm{a}_{t}\sim\pi(\cdot\mid\bm{s}_{t})}\nonumber\\
&&\left[\gamma^{t}r(\bm{s}_{t},\bm{a}_{t},\bm{s}_{t+1})\right],\label{S2E1}
\end{eqnarray}
by finding an optimal policy $\pi_{\bm{\theta}^{\pi}}^{*}(\bm{s})$ parameterized by $\bm{\theta}^{\pi}$.

In this paper, we use LLMs to improve the sample efficiency of the TD3 algorithm \cite{fujimoto2018addressing}. TD3 seeks to maximize the objective \eqref{S2E1} by finding a deterministic policy $\pi_{\bm{\theta}^{\pi}}^{*}(\bm{s}):\mathcal{S}\mapsto\mathcal{A}$. Compared with DDPG \cite{lillicrap2015continuous}, TD3 solves the overestimation bias of the $Q$-value function by employing a variant strategy of double Q-learning \cite{hasselt2010double, van2016deep}. As an off-policy RL algorithm, TD3 updates two critics $Q_{\bm{\theta}_{i}^{Q}}$, $i=1,2$, by solving
\begin{equation}
\min_{\bm{\theta}_{i}^{Q}}\frac{1}{N}\sum_{n=1}^{N}(y_{n}-Q_{\bm{\theta}_{i}^{Q}}(\bm{s}_{n},\bm{a}_{n}))^{2},\quad i=1,2,
\label{S2E2}
\end{equation}
where $N$ is the batch size of data sampling from a replay buffer and $y_{n}$ is the critic target constructed by
\begin{equation}
y_{n}\triangleq r_{n}+\gamma\min_{i\in\{1,2\}}Q_{\bar{\bm{\theta}}_{i}^{Q}}(\bm{s}_{n}^{\prime},\pi_{\bar{\bm{\theta}}^{\pi}}(\bm{s}_{n}^{\prime})+\bm{\epsilon}_{n}),
\end{equation}
with $Q_{\bar{\bm{\theta}}_{i}^{Q}}$, $i=1,2$, and $\pi_{\bar{\bm{\theta}}^{\pi}}$ being the target networks that slowly track the learned networks and $\bm{\epsilon}_{n}$ is a noise that regularizes the value function to be smooth. The actor of TD3 is updated by employing the deterministic policy gradient \cite{silver2014deterministic}
\begin{equation}
\nabla_{\bm{\theta}^{\pi}} J(\bm{\theta}^{\pi})=\frac{1}{N}\sum_{n=1}^{N}\left.\nabla_{\bm{a}}Q_{\bm{\theta}_{1}^{Q}}(\bm{s}_{n},\bm{a})\right|_{\bm{a}=\pi_{\bm{\theta}^{\pi}}(\bm{s}_{n})}\nabla_{\bm{\theta}^{\pi}}\pi_{\bm{\theta}^{\pi}}(\bm{s}_{n}).
\end{equation}
{\color{black}In TD3, the update frequency of the critic is $d$ times the update frequency of the actor.} 

\subsection{Problem Descriptions}
\label{S22}
In this paper, we consider the RL problems on a set of tasks with the Franka Emika Panda robot. Our approach, aligned with prior research \cite{nair2018overcoming, gallouedec2021panda, andrychowicz2017hindsight}, operates under the assumption that only a basic movement controller for the robot is available. Specifically, we assume that the 3-dimensional gripper position can change within the range of [$-$0.05 m, 0.05 m] and the 3-dimensional Euler angles can change within the ranges of [$-$0.15 rad, 0.15 rad] if permitted. This setup excludes the use of a high-level motion planner for long-distance gripper movements. The robot tasks involve picking and placing, sliding or pushing objects, etc. These tasks are challenging due to their long time horizons and the sparse reward structure: 0 for incomplete tasks and 1 for complete ones.

Our goal is to utilize the inherent knowledge of LLMs to accelerate the RL processes in terms of improving sample efficiency. The existing works have shown LLMs' capability in generating reward functions for robot RL \cite{yu2023language, xie2023text2reward, ma2023eureka}, and their user-friendly nature with natural language communication. However, effectively using LLMs to reduce the sample complexity in RL remains an unexplored area.

\section{Methods of RLingua}
\label{S3}
\begin{figure*}[!t]
\centering
\includegraphics[width=\textwidth]{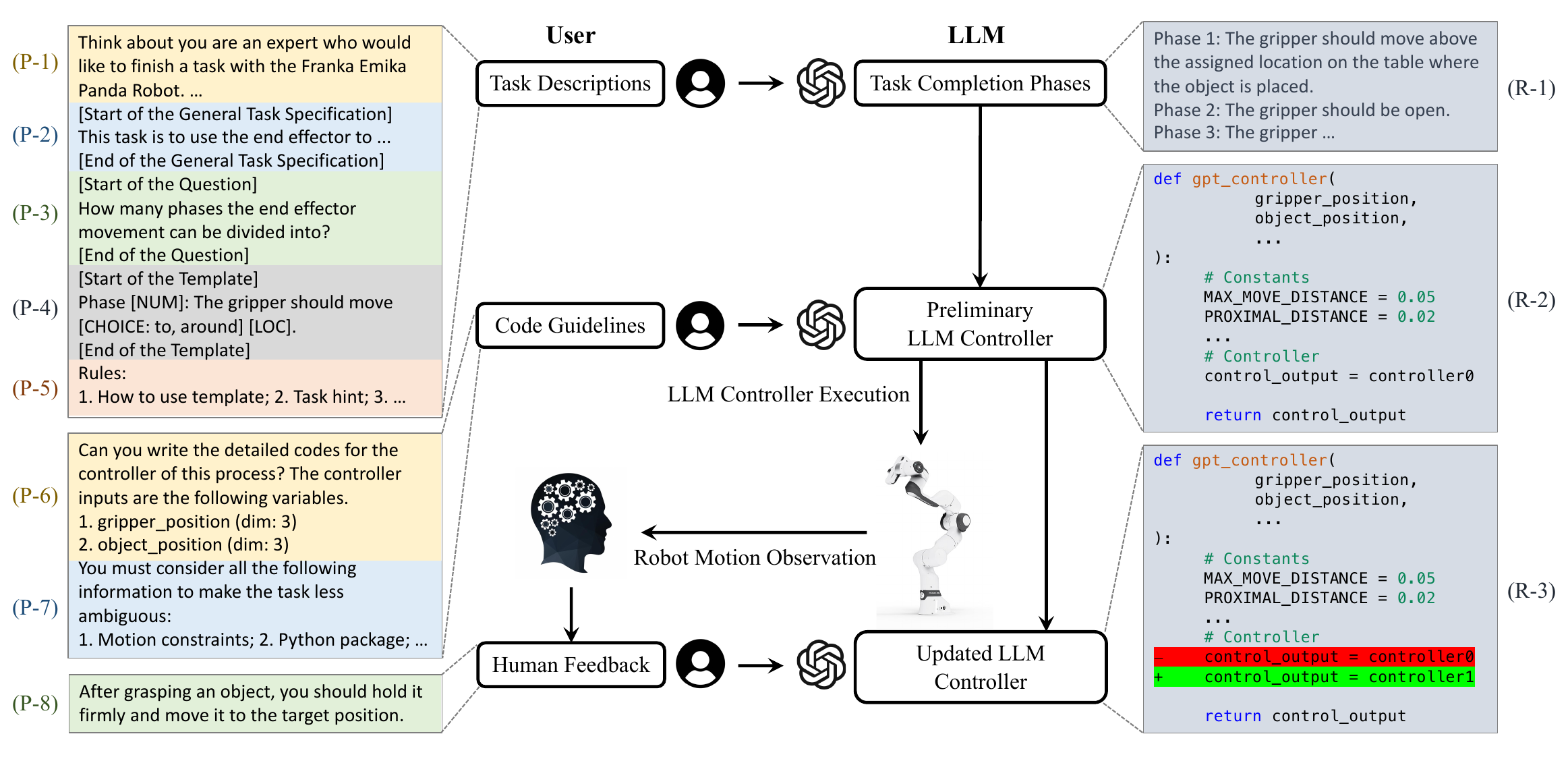}
\caption{The framework of prompt design with human feedback. The task descriptions and code guidelines are prompted in sequence. The human feedback is provided after observing the preliminary LLM controller execution process on the robot.}
\label{Fig3}
\end{figure*}

This section details the methods of RLingua, which bridges LLMs with RL using an LLM-generated controller. The overall framework is depicted in Fig. \ref{Fig5} with specific prompt design processes shown in Figs. \ref{Fig3} and \ref{Fig4}. RLingua's methodology consists of the following two primary steps.

\begin{itemize}
\item [i)] \textit{LLM Prompt Design:} Designing prompts to extract the internal knowledge of LLMs, forming a preliminary controller for the task at hand. The obtained robot controller is rule-based and written in Python. It is typically imperfect due to the fact that LLMs cannot perfectly capture the task dynamics and robot motion limits. 
\item [ii)] \textit{RL with LLM-Generated Controller:} Utilizing the LLM-generated imperfect controller to generate samples that guide the explorations of RL, with the aim of reducing the required number of samples and improving the performance of the RL agent. 
\end{itemize}

\subsection{LLM Prompt Design}
\label{S31}
In this section, we will show the LLM-prompt design procedures for robot tasks. In RLingua, we extract the knowledge of LLMs in the form of a Python rule-based controller \cite{xu2022systematic, bubeck2023sparks}. This controller is typically imperfect since it, for example, cannot capture the robot task dynamics or may generate an infeasible path. We propose two types of prompt design: prompt design with human feedback and prompt design with a code template.

\subsubsection{Prompt Design with Human Feedback}
\label{S311}
In order to obtain a preliminary controller from LLMs, we need to design a prompt that is concise but includes all the necessary information. This section provides a prompt design method by asking LLMs to first divide a long-horizon task into different elementary phases and then generate a rule-based controller in Python. In addition, we can improve the controller by implementing it on the robot, making observations of the performance, and providing feedback from humans, as shown in Fig. \ref{Fig3}. 

There are three components in the prompts to elicit the task-specific rule-based robot controller from LLMs: 1) task descriptions, 2) code guidelines, and 3) human feedback. The task descriptions and code guidelines are executed in sequence in an open-loop manner, while human feedback is provided based on the performance of controller execution. Compared with combining task descriptions and code guidelines as one prompt, the two-step prompting method makes the LLMs more focused on subquestions, which we design to guide the LLMs toward the final controller code generation. 

{\textit{Task Descriptions:}} 
The task descriptions include the basic settings of the robot and the task. In our prompts, we provide the following components, of which the indexes are marked in Fig. \ref{Fig3}. 
\begin{itemize}[leftmargin=0.9cm]
\item [(P-1)] Robot information: This part describes the robot type, control variables, and robot constraints.
\item [(P-2)] General task descriptions: This component describes the task the robot needs to complete and the detailed information about the objects involved in the task. 
\item [(P-3)] Question description: This step describes the question the LLMs should answer. 
\item [(P-4)] Response template: This standardizes the LLM's response. 
\item [(P-5)] Rules: The rules help LLMs better understand the question. 
\end{itemize}

The task descriptions provide a standard framework that a non-expert can generalize to different robot tasks. In detail, the components (P-1), (P-3), and (P-4) are pre-designed by experts and are fixed across all the tasks with the same robot. The component (P-2) is specific to tasks and should be given by any non-expert. The users can provide hints about implementing a robot task in (P-5).

After prompting the task descriptions to LLMs, we can obtain the phase descriptions of the robot to complete the task. This phase information will then be used in the next stage, together with code guidelines, to direct LLMs in generating code for LLM controller.

{\textit{Code Guidelines:}} We design code guidelines as the further prompt so that the LLMs can output a preliminary task controller. The code guidelines include the following components. 

\begin{itemize}[leftmargin=0.9cm]
\item [(P-6)] Control inputs: In order for LLMs to generate an executable and compatible robot controller in Python, we clearly define control inputs, robot, and environmental states. The LLMs may only use a subset of these variables as inputs at their discretion. Thus, we can let an expert provide all candidate control inputs in advance and let LLMs decide which one should be used. 
\item [(P-7)] Rules: Hints for LLMs to generate a rule-based controller in Python, such as the usages of Python packages.
\end{itemize}

{\textit{Human Feedback:}} 
Utilizing the provided coding guidelines, a preliminary Python-based robot controller can be generated by LLMs. Our experiments reveal that the controllers created in this manner occasionally fail to accomplish the specified robotic tasks. For instance, under such a controller, the robot's gripper might not securely grasp an object when tasked with moving it from one location to another. In such scenarios, we can submit feedback (P-8) to the LLMs, enabling them to revise and enhance the performance of the initially generated controller. Unlike the feedback process described in \cite{wake2023chatgpt}, which necessitates user expertise in task planning, RLingua assumes that users will provide feedback at a non-expert level, using natural language. {\color{black}Thus, there will not be a steep learning curve for end users to prompt the method.}
\begin{figure*}[!t]
\centering
\includegraphics[width=\textwidth]{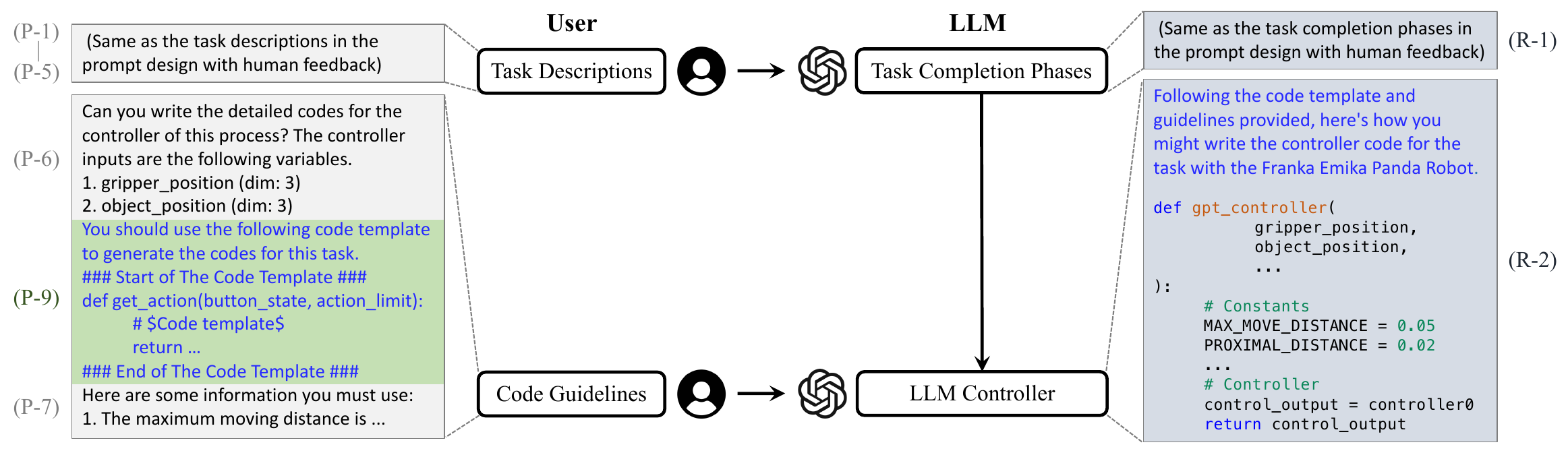}
\caption{The framework of prompt design with a code template}
\label{Fig4}
\end{figure*}

\subsubsection{Prompt Design With A Code Template}
\label{S312}
The controller code generation with human feedback in Section \ref{S311} is suitable for robot tasks with flexible dynamics but simple control logics. For long-horizon tasks with high dimensional state and action spaces, the LLMs currently available may fail to generate an executable Python controller with only human feedback. To solve this issue, we can provide a code template as a part of the code guidelines and ask the LLMs to follow the template when generating controller codes. Fig. \ref{Fig4} shows the framework for the prompt design with a code template. Similar to the prompt design with human feedback in Section \ref{S311}, we prompt the LLMs to generate task completion phases by inputting task descriptions (P-1)--(P-5) before code guidelines. The details of task descriptions and task completion phases in Fig. \ref{Fig4} are the same as those in Fig. \ref{Fig3}. Different from the prompts in \cite{huang2022language, ahn2022can, huang2022inner, zhang2023bootstrap} that assume the availability of robot primitives, the code template (P-9) in our prompts is based on the assumption that the robot can only execute basic motions. Thus, the LLMs need to figure out how to complete a long-horizon task by executing the basic motions cumulatively.

We provide two examples of prompt designs with human feedback and a code template, respectively, in Appendix \ref{SA}\footnote{The appendix sections can be found at \url{https://rlingua.github.io}.}. 
	
\subsection{Reinforcement Learning With the LLM Controller}
\label{S32}
This section shows the method of using the LLM-generated controller to improve the sample efficiency of RL, as presented in Fig. \ref{Fig5}. We can use the LLM controller $\pi_{\mathrm{LLM}}$ to generate data by interacting with the environment. The sampled data are stored in a replay buffer for further use in policy learning \cite{nair2018overcoming}. We maintain two replay buffers $\mathcal{R}_{\mathrm{LLM}}$ and $\mathcal{R}_{\mathrm{RL}}$ that separately store the data generated by the LLM controller and the policy being trained $\pi_{\bm{\theta}^{\pi}}$. The actor loss is constructed as
\begin{equation}
\ell_{\mathrm{actor}}(\bm{\theta}^{\pi})\triangleq\ell_{\mathrm{RL}}(\bm{\theta}^{\pi})+\lambda_{\mathrm{IM}}\ell_{\mathrm{IM}}(\bm{\theta}^{\pi}),
\end{equation}
with
\begin{eqnarray}
\ell_{\mathrm{RL}}(\bm{\theta}^{\pi})&\triangleq&\frac{1}{N}\sum_{\{\bm{s}_{n}\}_{n=1}^{N}\in\mathcal{R}_{\mathrm{RL}}}Q_{\bm{\theta}_{1}^{Q}}(\bm{s}_{n},\pi_{\bm{\theta}^{\pi}}(\bm{s}_{n})),\\
\ell_{\mathrm{LLM}}(\bm{\theta}^{\pi})&\triangleq&\frac{1}{N}\sum_{\{\bm{s}_{n},\bm{a}_{n}\}_{n=1}^{N}\in\mathcal{R}_{\mathrm{LLM}}}(\pi_{\bm{\theta}^{\pi}}(\bm{s}_{n})-\bm{a}_{n})^{2},
\end{eqnarray}
where $\lambda_{\mathrm{IM}}$ is a hyperparameter controlling the relative weight of imitation learning loss and $Q_{\bm{\theta}_{1}^{Q}}$ is the Q-value function in TD3 updated via \eqref{S2E2}, and $N$ denotes the batch size. We can regard $\ell_{\mathrm{LLM}}$ as a regularization to the standard actor loss. 

We consider the online RL in the framework of RLingua, of which the data are collected during the training process. In RLingua, the LLM controller and the policy being trained are used to make rollouts in a combined manner. In detail, with the current state $\bm{s}_{t}$, we sample the next action $\bm{s}_{t}$ from the LLM-policy $\pi_{\mathrm{LLM}}$ with a probability of $p^{\mathrm{LLM}}$ and from the training policy $\pi_{\bm{\theta}^{\pi}}$ with a probability of $1-p^{\mathrm{LLM}}$. This is different from the method in \cite{nair2018overcoming} since the LLM controller guides the data collection process in the {\textsc{DAgger}} manner, which has been proved to be helpful in reducing the compounding loss of imitation learning due to distribution shift \cite{ross2011reduction}. Since the rule-based LLM-generated controller is typically imperfect, we anneal the probability $p^{\mathrm{LLM}}$ as the learning process continues. In RLingua, the value of $p^{\mathrm{LLM}}$ decays exponentially as $p^{\mathrm{LLM}}_{k+1}=\lambda_{\mathrm{annl}}^{k}\cdot p^{\mathrm{LLM}}_{k}$, where $k$ denotes the number of environment steps since the training process starts. In Section \ref{S4}, we will show that this annealing strategy is effective and the imperfect of LLM-generated policy will be overcome by RLingua. The pseudo-code of RLingua is shown in Algorithm \ref{Alg1} in Appendix \ref{SA0}.

\section{Experiments}
\label{S4}
In this section, we will demonstrate the effectiveness of RLingua by both simulated and real robot tasks. Sections \ref{S41} and \ref{S42} show the simulation results on some tasks in the {\texttt{panda\_gym}} \cite{gallouedec2021panda} and {\texttt{RLBench}} environments \cite{james2020rlbench}, respectively. Section \ref{S43} presents the results on real robot tasks by Sim2Real. The robot in all these tasks is the Franka Emika Panda robot and the LLM that we use is the OpenAI GPT-4 (\url{https://openai.com/gpt-4}). 

\subsection{Simulations in the \texttt{panda\_gym} Environment}
\label{S41}
This section shows that RLingua is effective in reducing the sample complexity of RL for tasks involving dynamic interactions in {\texttt{panda\_gym}} \cite{gallouedec2021panda}. We note that the LLM-generated controllers are imperfect on some of the tasks due to the fact that the task dynamics are hard to incorporate in the LLM-generated codes. 

\begin{figure*}
\centering
\includegraphics[width=0.815\textwidth]{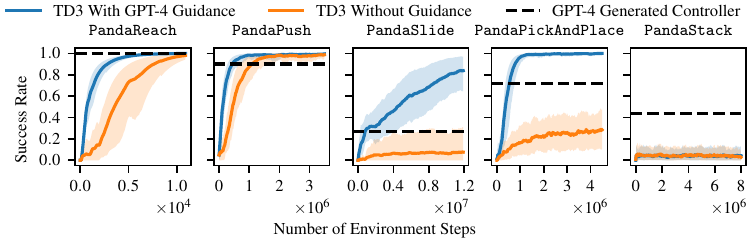}
\caption{\color{black}The success rates of different tasks in {\texttt{panda\_gym}} with respect to numbers of environment samples. The solid line is the mean success rate and the shaded region represents the minimum and maximum success rates, both evaluated with four different random seeds. The exponential moving average with a smoothing factor of 0.95 is applied to all curves.}
\label{Fig6}
\end{figure*}

We consider {\color{black}five} goal-oriented tasks in {\texttt{panda\_gym}} \cite{gallouedec2021panda}, which are (a) \texttt{PandaReach}, (b) \texttt{PandaPush}, (c) \texttt{PandaPickAndPlace}, (d) \texttt{PandaSlide}, {\color{black}and (e) \texttt{PandaStack}}. The detailed task descriptions and visualizations can be found in Appendix \ref{SCB}. In all {\color{black}five} tasks, the observations consist of the position and speed of the gripper. If a cube is involved in the task, then its position, speed, orientation in terms of Euler angles\footnote{The sequence of axes for rotations is $x$-$y$-$z$, with respect to the extrinsic frame.}, and rotational speed are included in the observations. The initial and target positions of the cube are randomly selected in different episodes. The controlled variables, i.e., the variables in the action space, are the 3-dimensional gripper movement subject to the constraints in Section \ref{S22}. In the \texttt{PandaPickAndPlace} and {\color{black}\texttt{PandaStack}} tasks, the indicator of whether the gripper is open is an additional action variable. 

The friction between the cube and table is small in the tasks of {\texttt{panda\_gym}}. In our experiment, we find that this will add to task difficulties. For example, in the \texttt{PickAndPlace} task, if the gripper cannot grasp the cube firmly on its first try, then the cube may slide away due to colliding with the gripper. The gripper may never be able to pick the cube if the cube slides beyond the gripper's reachable region. These sensitive task dynamics are the primary reason why the rule-based controllers generated by GPT-4 sometimes fail the tasks \texttt{PandaPush}, \texttt{PandaSlide}, and \texttt{PandaPickAndPlace}. For both the RLingua with TD3 and standard TD3 algorithm, the hindsight experience replay (HER) \cite{andrychowicz2017hindsight} are used in these goal-oriented tasks. Note that the five robot tasks in \texttt{panda\_gym} have simple control logic but different dynamics types, including pushing, sliding, picking and placing. The prompt design for these tasks uses the prompt design with human feedback in Section \ref{S311}. {\color{black} During the experiments, we found that the LLM-generated codes are not sensitive to specific prompt designs, as long as the task descriptions and code guidelines are completely stated and human feedback is appropriately provided.}

The \texttt{PandaReach} is a task that does not involve dynamic interactions. In this simple task, the GPT-4 generated controller realizes a success rate of 1. Fig. \ref{Fig6} shows that the success rate of this task with RLingua can match that of the GPT-4 generated controller. Compared with the TD3 algorithm without GPT-4 guidance, RLingua uses less than 25\% environment steps to reach a success rate of 0.8. The other four tasks involve the interactions between the robot gripper and cube, causing the GPT-4 generated rule-based controllers to be imperfect due to the sensitive dynamics. Fig. \ref{Fig6} demonstrates that RLingua can outperform or match the GPT-4 generated controllers in all tasks {\color{black}excluding \texttt{PandaStack}}. For the \texttt{PandaPush} task, RLingua needs less than 50\% environment steps to reach a success rate of 0.8. For the tasks \texttt{PandaSlide} and \texttt{PandaPickAndPlace}, RLingua realizes significantly larger success rates with the same number of environment steps than the standard TD3 algorithms, of which the success rates improve slowly with millions of environment samples. The results indicate that RLingua can leverage the prior knowledge of GPT-4 to improve the sample efficiency and overcome the imperfect of GPT-4 generated controllers with the adoption of RL. {\color{black}It is noteworthy that both the standard RL and RLingua nearly fail the \texttt{PandaStack} task, even though the GPT-4 generated controller achieves a success rate of 0.44. We hypothesize that this failure is attributable to the task's long horizon, which significantly increases reward sparseness for RL and complicates learning for imitation learning. However, we can split the \texttt{PandaStack} task into two \texttt{PandaPickAndPlace} tasks which have a success rate of 1 after training with RLingua at the current stage.}

\subsection{Simulations in the \texttt{RLBench} Environment}
\label{S42}
This section demonstrates that RLingua can leverage the LLM-generated robot controllers to solve robot tasks in \texttt{RLBench} \cite{james2020rlbench}, while the standard RL fails. {\color{black}We sample 12 tasks considering the task diversity, the capacity of GPT-4, and the task brittleness to RL explorations. The detailed selection criteria and the visualization of the selected tasks are stated in Appendix \ref{SC}.} Unlike the tasks in \texttt{panda\_gym}, the gripper orientation is considered part of the task state in \texttt{RLBench}.

\begin{figure*}
\centering
\includegraphics[width=\textwidth]{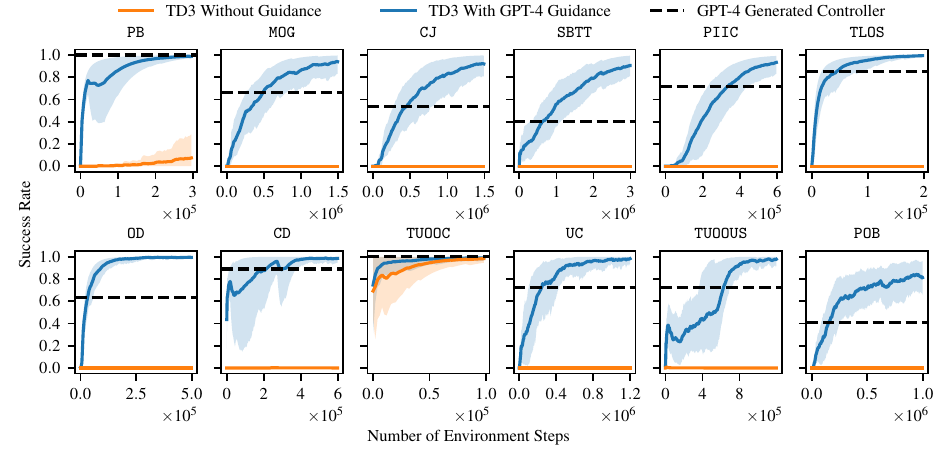}
\caption{\color{black}The success rates of different tasks in {\texttt{RLBench}} with respect to numbers of environment samples. The solid line is the mean success rate and the shaded region represents the minimum and maximum success rates, both evaluated with four different random seeds. The exponential moving average with a smoothing factor of 0.95 is applied to all curves.}
\label{Fig7}
\end{figure*}

The positions of objects involved in the \texttt{RLBench} tasks are selected randomly in different episodes. The observations consist of the position and Euler angles of the robot gripper and involved objects. The actions consist of the 6-dimensional gripper displacement and rotation subject to the constraints in Section \ref{S22}, and one variable indicating whether the gripper is open or closed. The prompt design for these tasks uses a code template, as shown in Section \ref{S32}, since the robot controllers need to handle both gripper displacement and rotation, making it challenging for LLMs to generate a reasonable controller with only human feedback.

Fig. \ref{Fig7} shows the success rate of these tasks with respect to the number of environment samples. RLingua achieves success rates larger than 0.8 on all tasks, while the standard TD3 fails on 10 tasks completely with the same number of environment steps. The results demonstrate that RLingua, with the guidance of GPT-4 generated controllers, overcomes the problem of sparse rewards in these tasks and obtains more effective samples so as to achieve significantly improved success rates on the considered tasks.

In our experiments, we observe that the GPT-4 generated gripper path may be infeasible due to the motion constraints or the existence of obstacles. The results in Fig. \ref{Fig7} show that RLingua can overcome this path planning issue by RL and reach higher success rates than the GPT-4 generated controllers. 

\subsection{Real Robot Experiments}
\label{S43}
This section presents the Sim2Real experiments conducted using a real Franka Emika Panda robot for the task \texttt{put\_item\_in\_container} in \texttt{RLBench}. Utilizing the Sim2Real approach, the policies learned by RLingua demonstrate high transferability to real-world scenarios, particularly in robot pick and place tasks. The tasks involve the robot picking and placing one or more cubes into a container, a process during which the robot continuously receives real-time data from perceptual models. This allows the learned policies to predict the optimal actions at each step for the task at hand, ensuring robustness against external interference.

An Azure Kinect camera is used to detect the cube's position, enabling the extraction of the cube state. This setup is critical in tasks that involve altering the state of the target object and employing straightforward control logic, showcasing RLingua's adeptness in performing long-horizon, complex tasks. 

We validated the trained policy's performance on the following three different variants. 

\begin{itemize}
\item [(a)] Picking up a single cube and placing it in the container. This task demonstrates the basic pick-and-place ability of the robot, a fundamental aspect of the learned policies.
\item [(b)] Move the cube to various random locations multiple times with a stick while the gripper attempts to pick it up. This variant tests the policy's ability to adapt to changing conditions and unexpected movements, a key factor in real-world application.
\item [(c)] Picking different cubes and placing them in the container sequentially. In this complex and long-horizon task, we used cubes labeled with the letters ``B', ``A'', ``I'', ``D'', and ``U'', challenging the robot to identify and manipulate multiple targets.
\end{itemize}

\begin{figure}
\centering
\scalebox{0.79}{
\begin{minipage}{0.15\textwidth}
\includegraphics[width=\textwidth]{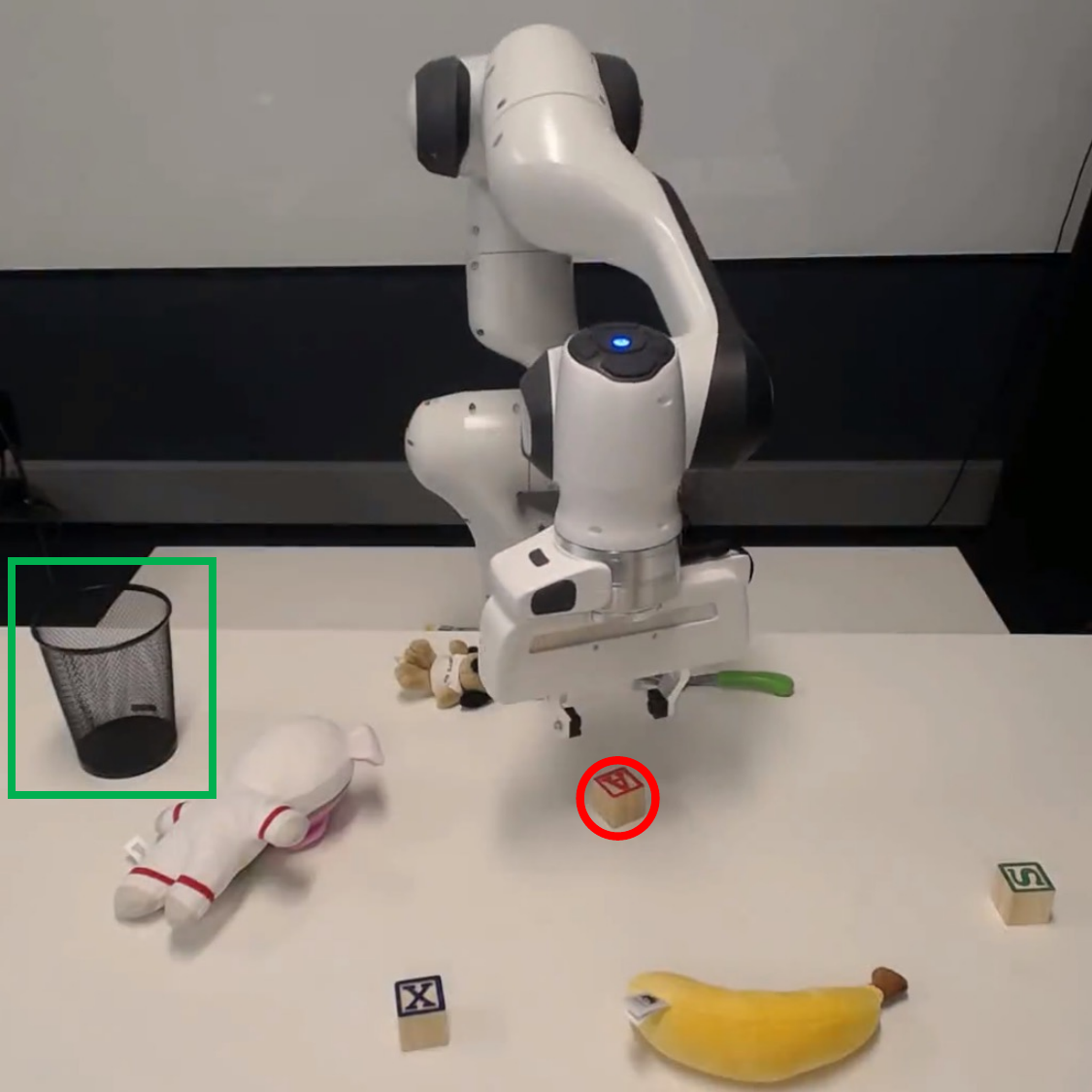}
\end{minipage}
\begin{minipage}[t]{0.015\textwidth}
\ding{212}
\end{minipage}
\begin{minipage}{0.15\textwidth}
\includegraphics[width=\textwidth]{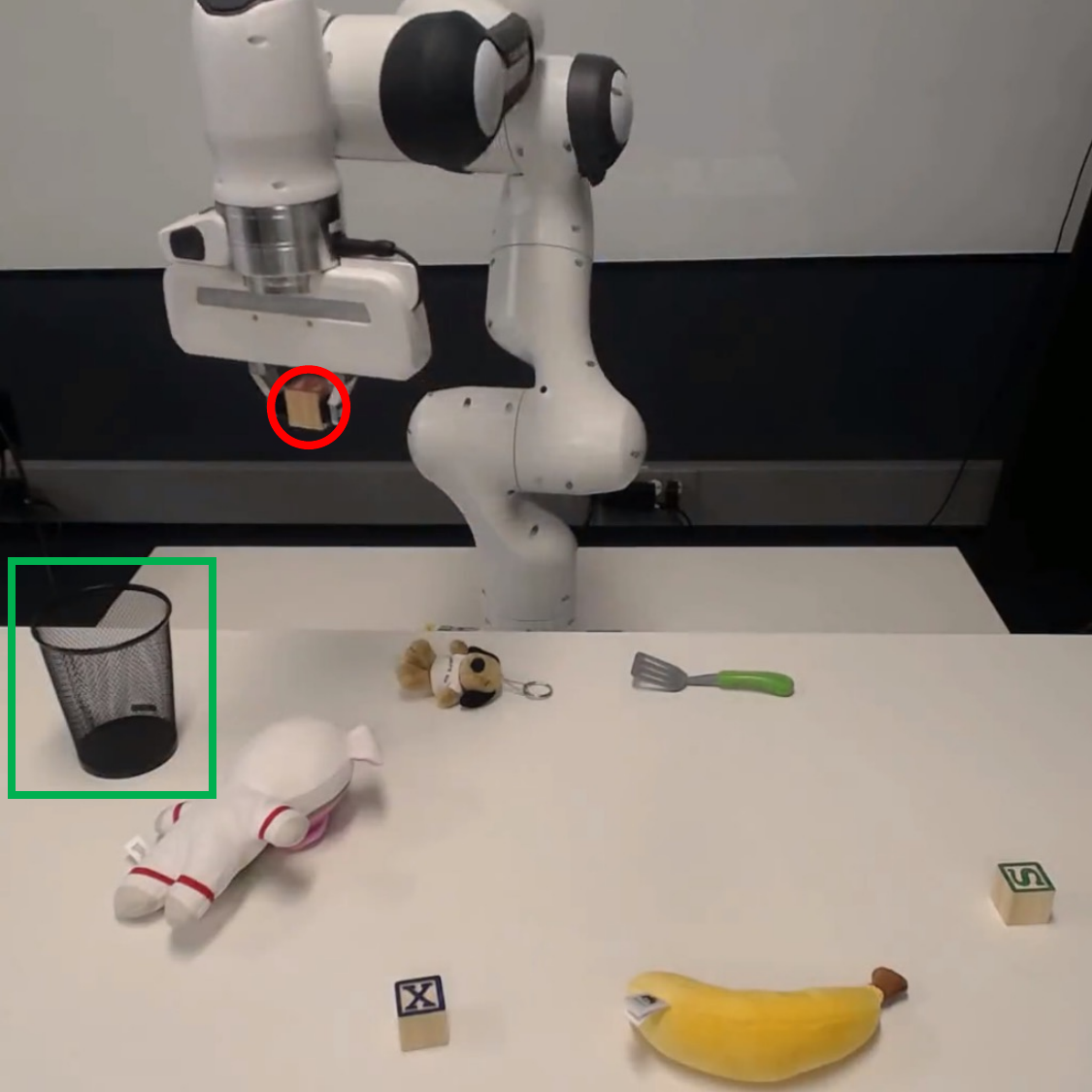}
\end{minipage}
\begin{minipage}{0.015\textwidth}
\ding{212}
\end{minipage}
\begin{minipage}{0.15\textwidth}
\includegraphics[width=\textwidth]{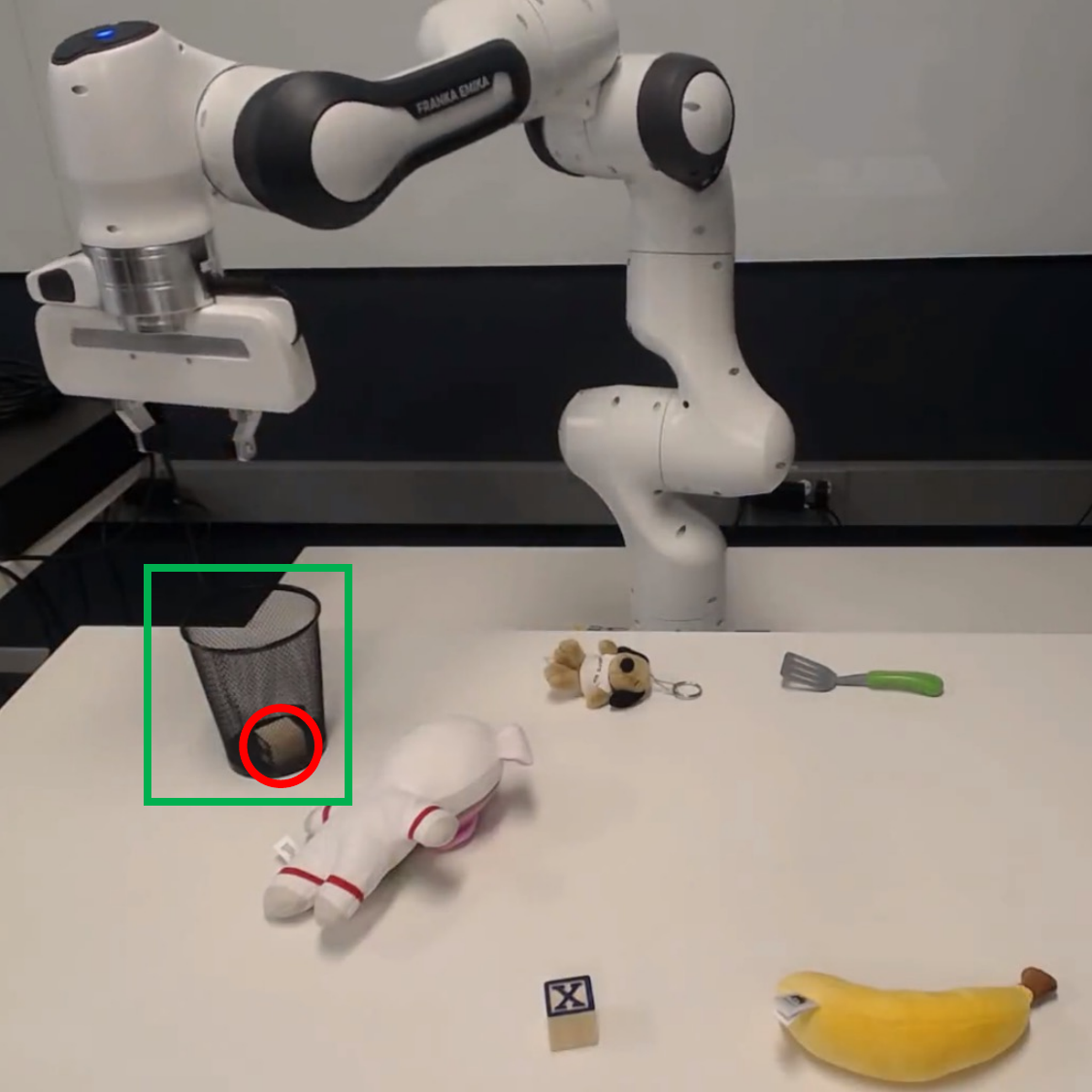}
\end{minipage}}

\vspace{0.1cm}
{\footnotesize(a) Single cube}
\vspace{0.1cm}

\scalebox{0.79}{
\begin{minipage}{0.15\textwidth}
\includegraphics[width=\textwidth]{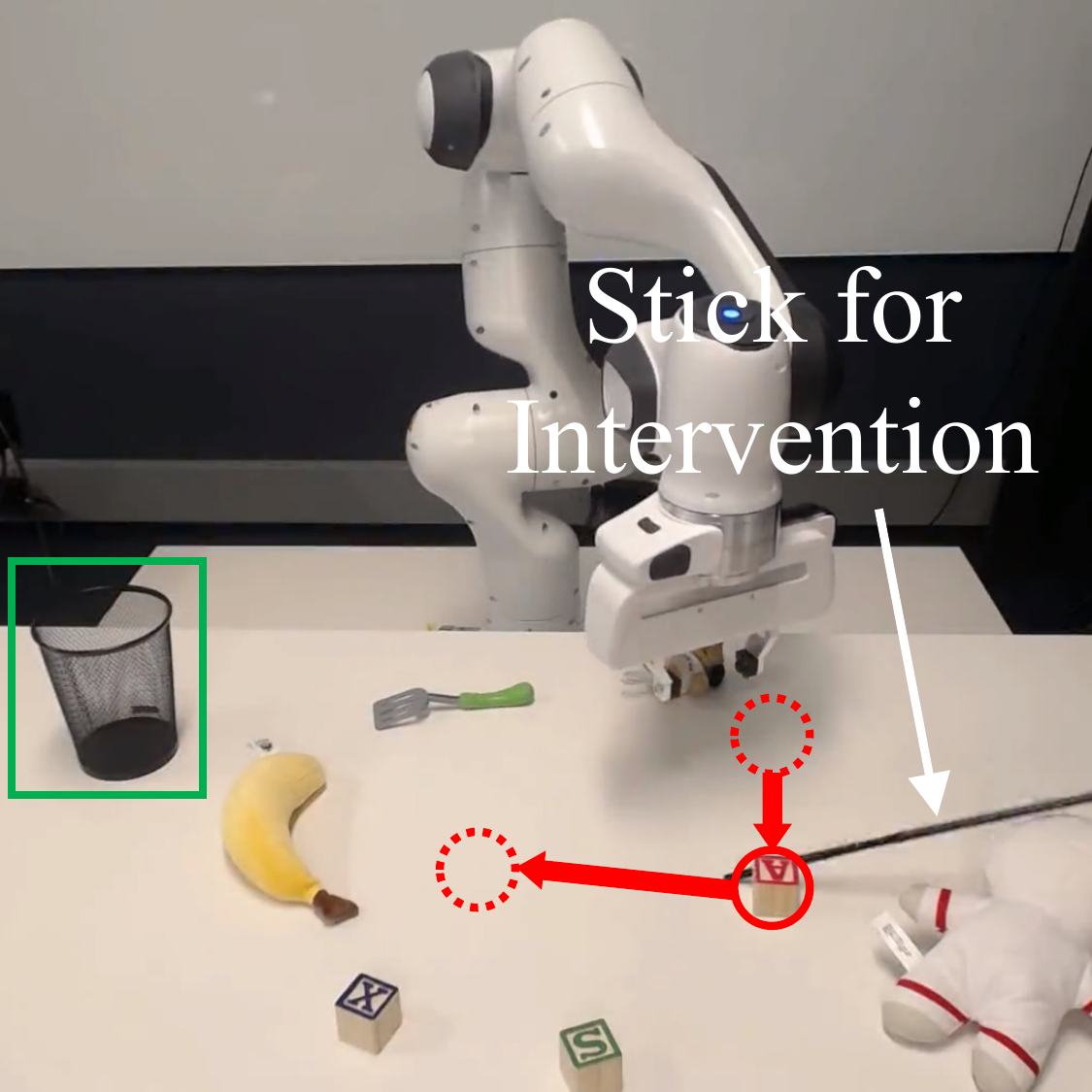}
\end{minipage}
\begin{minipage}[t]{0.015\textwidth}
\ding{212}
\end{minipage}
\begin{minipage}{0.15\textwidth}
\includegraphics[width=\textwidth]{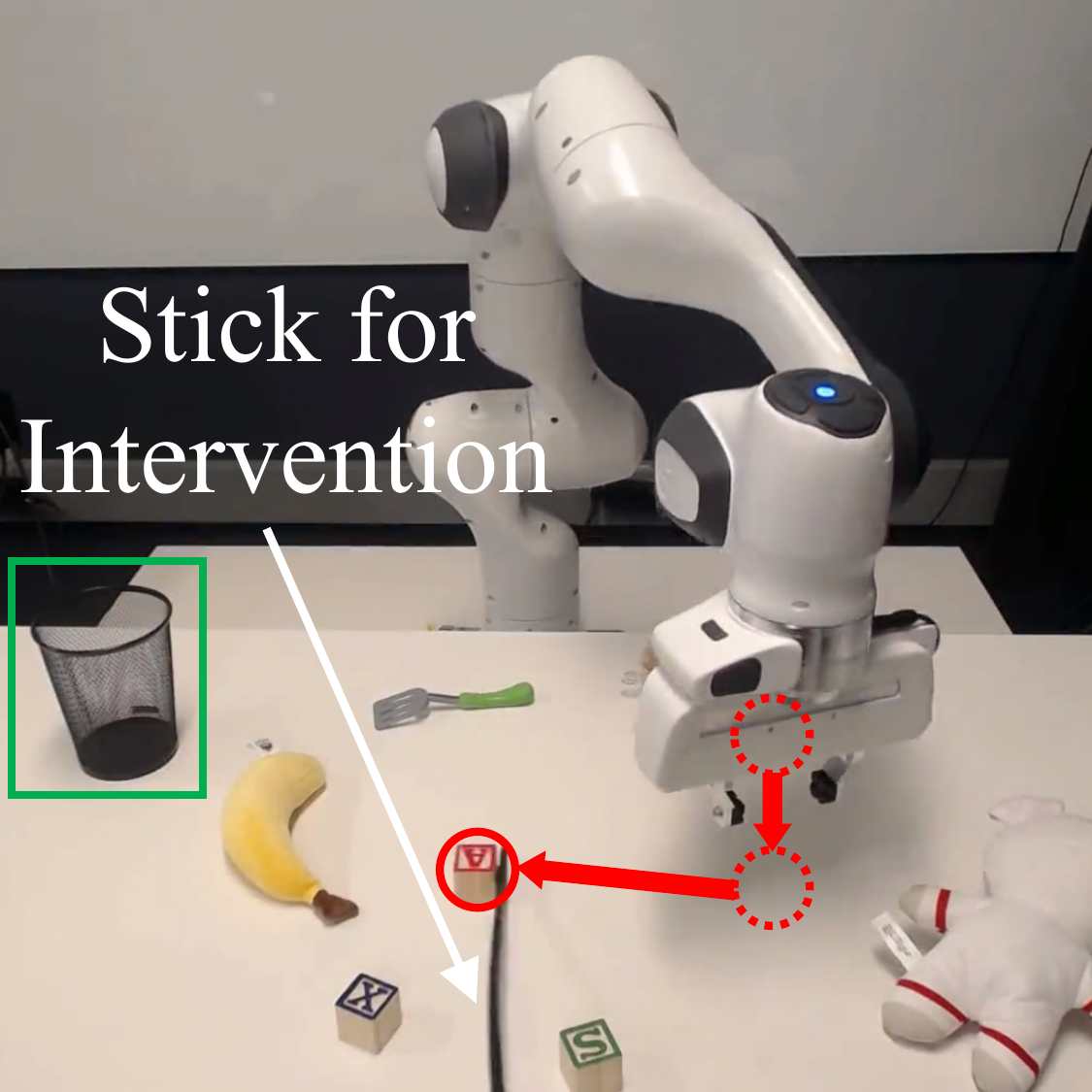} 
\end{minipage}
\begin{minipage}{0.015\textwidth}
\ding{212}
\end{minipage}
\begin{minipage}{0.15\textwidth}
\includegraphics[width=\textwidth]{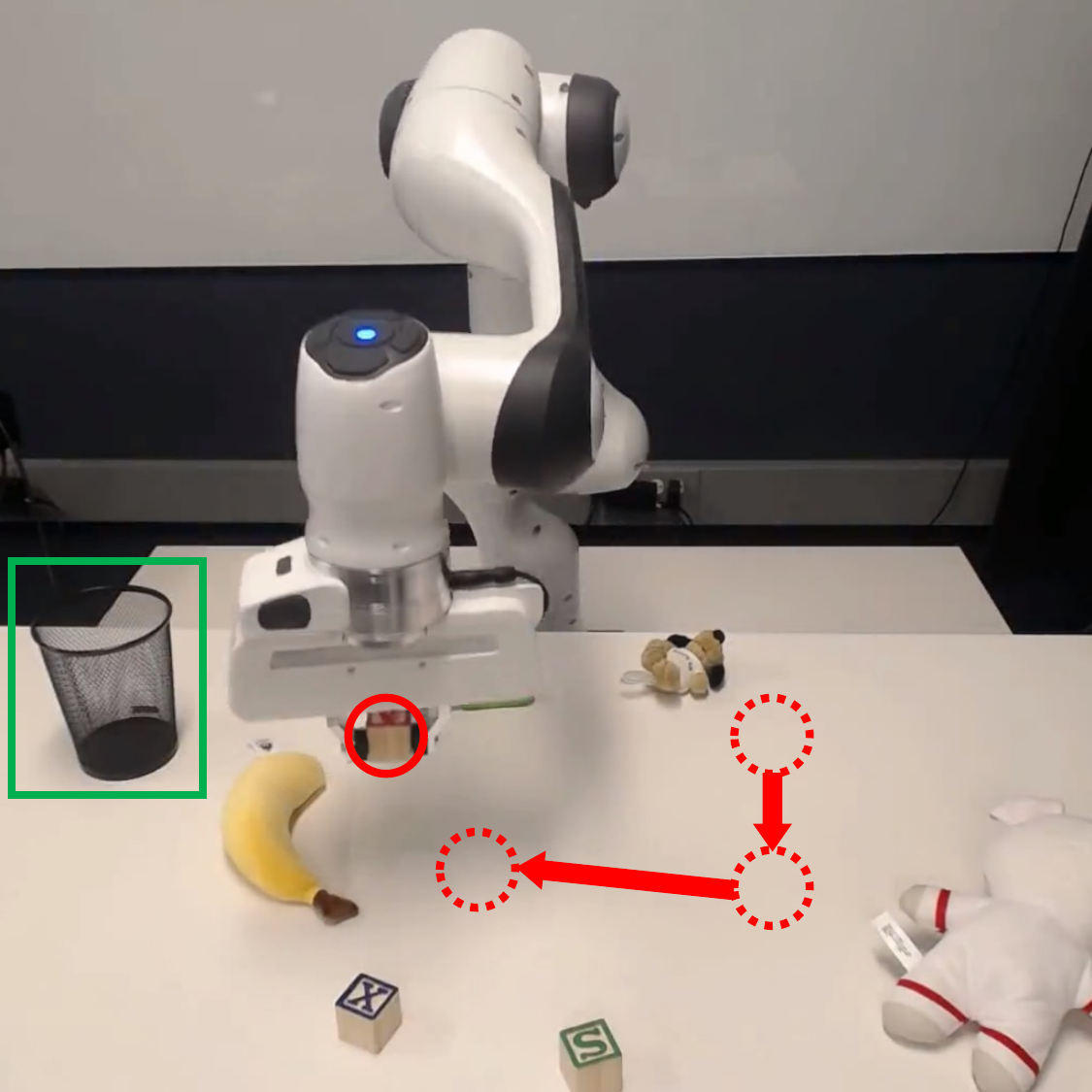}
\end{minipage}}

\vspace{0.1cm}
{\footnotesize(b) Single cube with interference}
\vspace{0.1cm}

\scalebox{0.79}{
\begin{minipage}{0.15\textwidth}
\includegraphics[width=\textwidth]{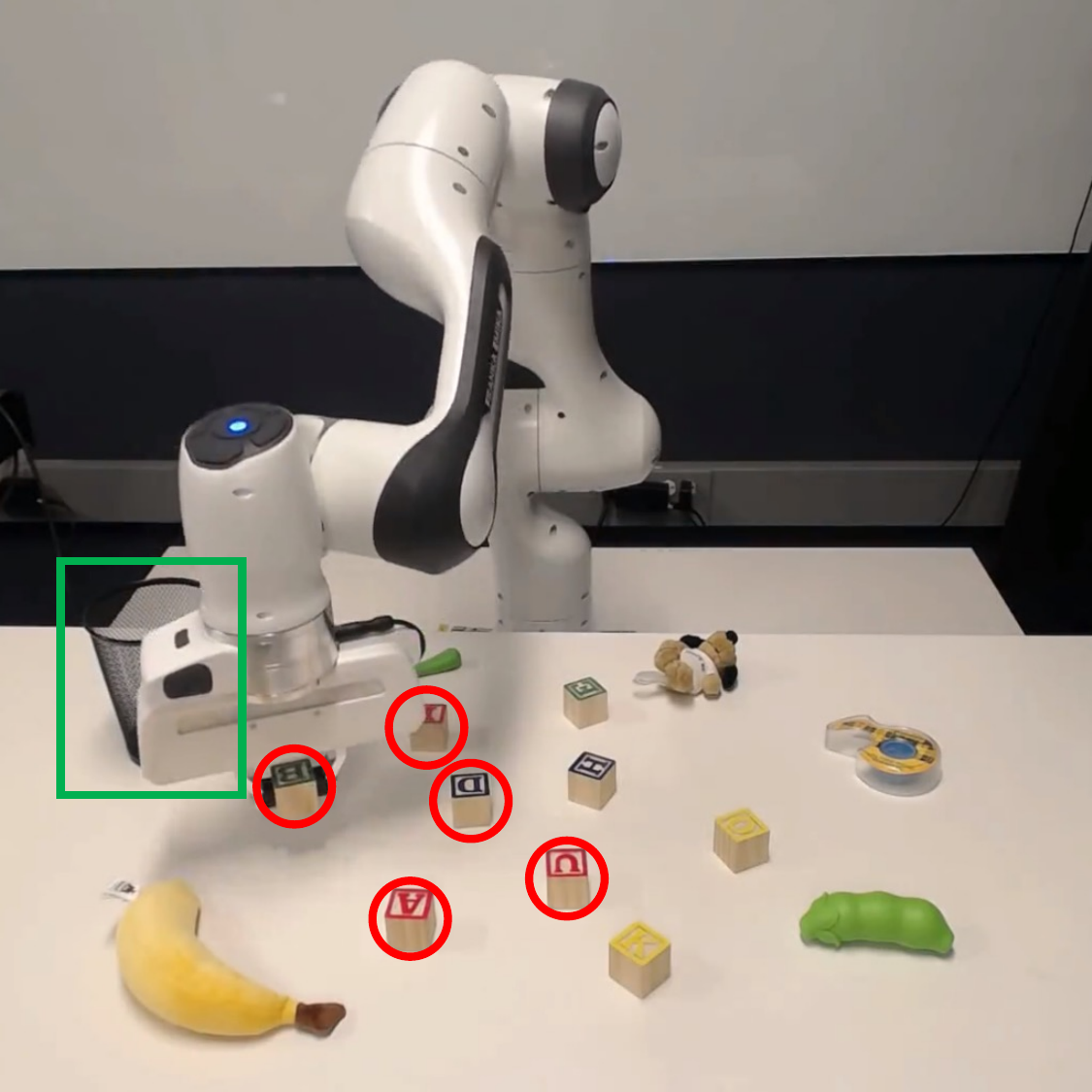}
\end{minipage}
\begin{minipage}[t]{0.015\textwidth}
\ding{212}
\end{minipage}
\begin{minipage}{0.15\textwidth}
\includegraphics[width=\textwidth]{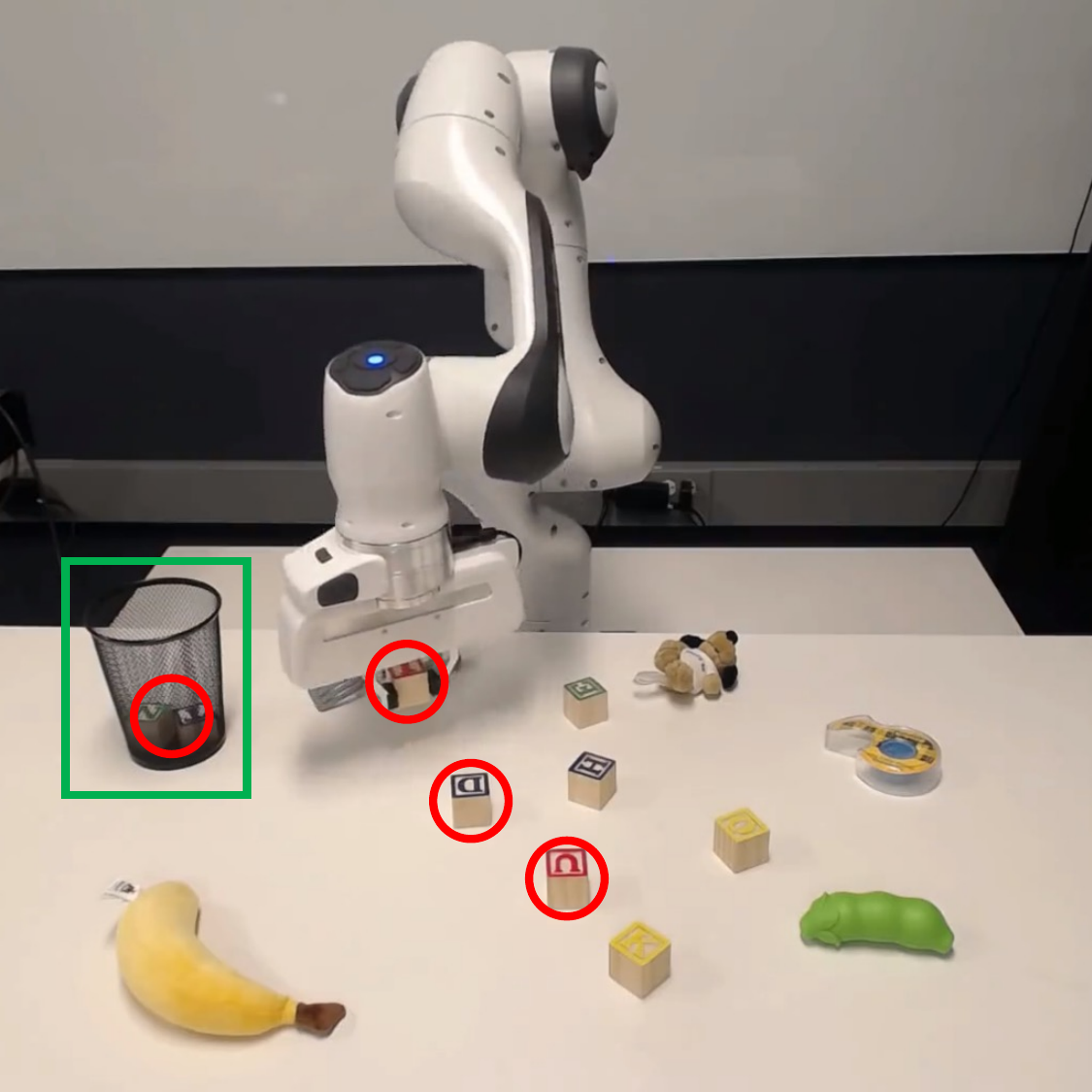} 
\end{minipage}
\begin{minipage}{0.015\textwidth}
\ding{212}
\end{minipage}
\begin{minipage}{0.15\textwidth}
\includegraphics[width=\textwidth]{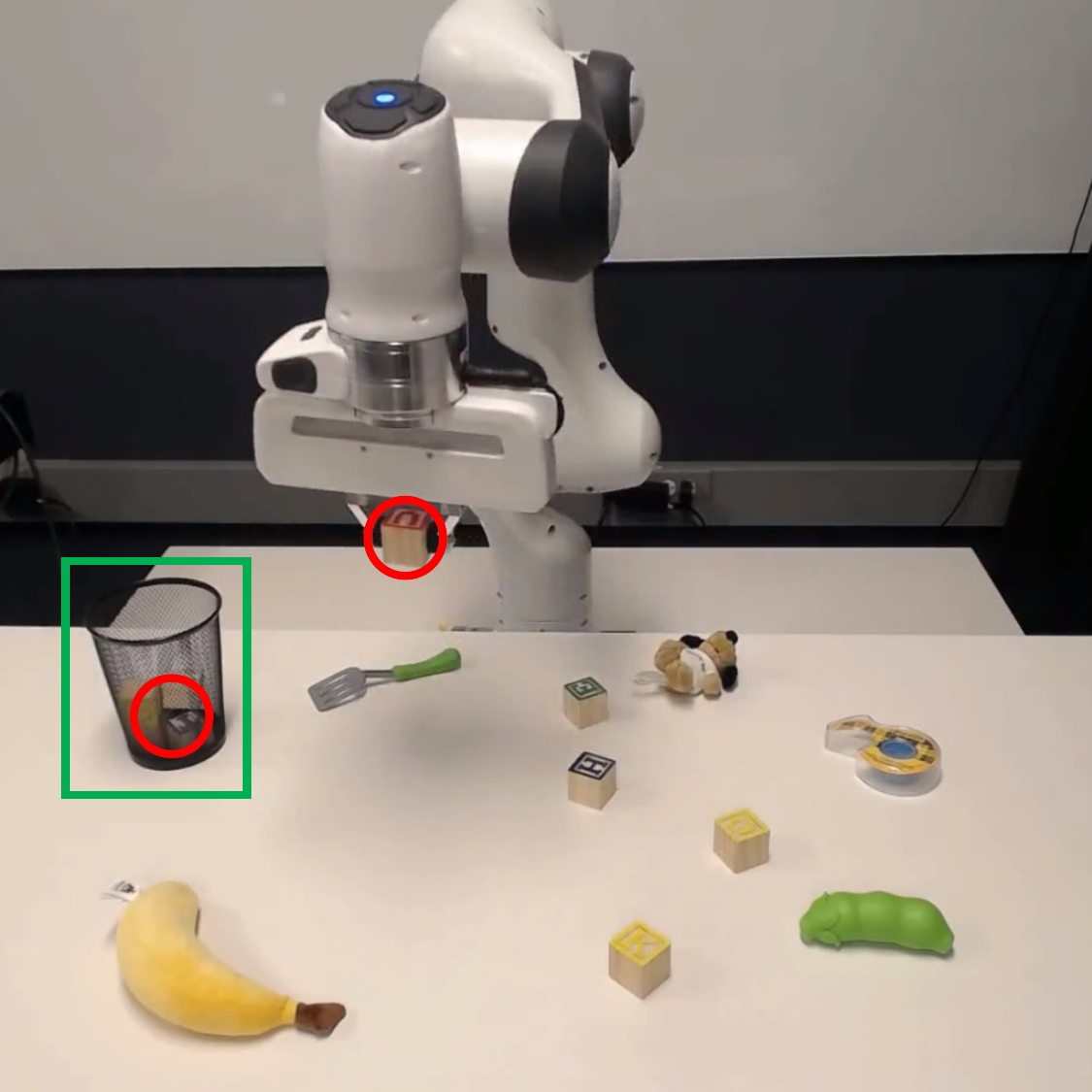}
\end{minipage}}

\vspace{0.1cm}
{\footnotesize(c) Multiple cubes with the letters ``B", ``A", ``I", ``D", and ``U" in order}
\vspace{0.1cm}

\caption{The real robot experiments}
\label{Fig8}
\end{figure}

Fig. \ref{Fig8} shows three screenshots from each task variant's implementation process. The successes of these tasks are indicative of the effective transfer of the learned policy in simulation by RLingua to real robot implementations. This achievement underscores the viability of the Sim2Real approach in bridging the gap between simulated training and practical application in our robotic systems.

\section{Conclusions}
In this paper, we propose the framework of RLingua that extracts the knowledge of LLMs about robot motions to rule-based controllers and further leverages the imperfect rule-based controllers to improve the sample efficiency of RL. From another perspective, RLingua can be regarded as using RL to improve the imperfect robot controllers directly generated by LLMs. We present two different prompt design procedures, i.e., the prompt design with human feedback and the prompt design with code template, to obtain the rule-based robot controllers from LLMs. In order to reduce the sample complexity of RL, we used the LLM-generated controllers to collect data during the online RL process. We modified the actor loss of the TD3 method and incorporated an imitation learning term as regularization so that the RL process is accelerated. 

We tested RLingua on some robot tasks with sparse rewards in \texttt{panda\_gym} and \texttt{RLingua}. The results demonstrated that RLingua can significantly improve the sample efficiency of RL algorithms and the obtained policies with \texttt{RLingua} outperform the LLM-generated controllers, which are imperfect due to the unknown environment dynamics and infeasible planned path. It is also worth noting that RLingua can solve the tasks in \texttt{RLBench} with high success rates, while the standard RL algorithm fails these tasks. The real robot experiments via Sim2Real show that the learned policy by RLingua can transferred to real robot tasks.

A limitation of RLingua is that the state information needs to be extracted manually during training. The future research direction, stimulated by the development and increased affordability of multi-modal LLMs \cite{yang2023dawn}, involves developing an end-to-end RLingua-like framework.

\bibliographystyle{IEEEtran}  
\bibliography{bibfile} 
	
\newpage
\onecolumn
\appendices
\section{Pseudo-code of RLingua With TD3}
\label{SA0}
The pseudo-code of RLingua with TD3 is shown in Algorithm \ref{Alg1}.
\begin{algorithm}[!ht]
\algorithmtag{A-1}
\caption{RLingua With TD3}
\label{Alg1}
\begin{algorithmic}[1]
\renewcommand{\algorithmicrequire}{\textbf{Input:}}
\renewcommand{\algorithmicensure}{\textbf{Output:}}
\REQUIRE The robot task description and optional feedback from a non-expert, a competent LLM, the initial probability $p^{\mathrm{LLM}}$ sampling an action from the LLM-generated policy and its decay rate $\lambda_{\mathrm{annl}}$, two empty replay buffers $\mathcal{R}_{\mathrm{LLM}}$ and $\mathcal{R}_{\mathrm{RL}}$, the data batch size $N$, the reward discount factor $\gamma$, the weight of imitation learning loss $\lambda_{\mathrm{IM}}$, and the small smoothing parameter $\tau$ for updating target networks
\ENSURE A robot controller $\pi_{\bm{\theta}^{\pi}}$ for the specified task\\
\textit{Initialization}: The neural network parameters $\bm{\theta}^{\pi}$, $\bm{\theta}_{1}^{Q}$, and $\bm{\theta}_{2}^{Q}$, and the target network parameters $\bar{\bm{\theta}}^{\pi}\gets\bm{\theta}^{\pi}$, $\bar{\bm{\theta}}_{1}^{Q}\gets\bm{\theta}_{1}^{Q}$, and $\bar{\bm{\theta}}_{2}^{Q}\gets\bm{\theta}_{2}^{Q}$
\STATE Based on the task description (and human feedback), generate a rule-based controller $\pi_{\mathrm{LLM}}$ from the LLM
\WHILE{not done}
\WHILE{sample collection required}
\STATE $\bm{a}\sim \alpha\cdot\pi_{\mathrm{LLM}}(\cdot\mid\bm{s})+(1-\alpha)\cdot\pi_{\mathrm{LLM}}(\cdot\mid\bm{s})$ with $\alpha \sim\mathrm{Bernoulli}(p^{\mathrm{LLM}})$
\STATE Observe the next state $\bm{s}^{\prime}$ and reward $r$ and store the tuple $(\bm{s},\bm{a},\bm{s}^{\prime},r)$ to either $\mathcal{R}_{\mathrm{LLM}}$ or $\mathcal{R}_{\mathrm{RL}}$, depending on whether $\bm{a}$ is sampled from $\pi_{\mathrm{LLM}}$ or $\pi_{\bm{\theta}^{\pi}}$
\STATE $p^{\mathrm{LLM}}\gets\lambda_{\mathrm{annl}}\cdot p^{\mathrm{LLM}}$
\ENDWHILE
\STATE Randomly sample $N$ data from $\mathcal{R}_{\mathrm{RL}}$ and update the critic parameters $\bm{\theta}_{i}^{Q}$, $i=1,2$, by minimizing $\frac{1}{N}\sum_{\{\bm{s}_{n},\bm{a}_{n}\}_{n=1}^{N}\in\mathcal{R}_{\mathrm{RL}}}(y_{n}-Q_{\bm{\theta}_{i}^{Q}}(\bm{s}_{n},\bm{a}_{n}))^{2}$, $i=1,2$, respectively, where $y_{n}\triangleq r_{n}+\gamma\min_{i\in\{1,2\}}Q_{\bar{\bm{\theta}}_{i}^{Q}}(\bm{s}_{n}^{\prime},\pi_{\bar{\bm{\theta}}^{\pi}}(\bm{s}_{n}^{\prime})+\bm{\epsilon}_{n})$ with $\bm{\epsilon}_{n}$ being a smoothing regularization noise
\IF{actor update required}
\STATE Randomly sample $N$ data from $\mathcal{R}_{\mathrm{RL}}$ and $\mathcal{R}_{\mathrm{LLM}}$, respectively, and update the actor parameter $\bm{\theta}^{\pi}$ by following the gradient $\frac{1}{N}\sum_{\{\bm{s}_{n},\bm{a}_{n}\}_{n=1}^{N}\in\mathcal{R}_{\mathrm{RL}}}\left.\nabla_{\bm{a}}Q_{\bm{\theta}_{1}^{Q}}(\bm{s}_{n},\bm{a})\right|_{\bm{a}=\pi_{\bm{\theta}^{\pi}}(\bm{s}_{n})}\times\nabla_{\bm{\theta}^{\pi}}\pi_{\bm{\theta}^{\pi}}(\bm{s}_{n})+\frac{2\lambda_{\mathrm{IM}}}{N}\sum_{\{\bm{s}_{n},\bm{a}_{n}\}_{n=1}^{N}\in\mathcal{R}_{\mathrm{LLM}}}(\pi_{\bm{\theta}^{\pi}}(\bm{s}_{n})-\bm{a}_{n})^{\mathrm{T}}\nabla_{\bm{\theta}^{\pi}}\pi_{\bm{\theta}^{\pi}}(\bm{s}_{n})$
\STATE Update target networks $\bar{\bm{\theta}}_{i}^{Q}\gets\tau\bm{\theta}_{i}^{Q}+(1-\tau)\bar{\bm{\theta}}_{i}^{Q}$,  $i=1,2$, and
$\bar{\bm{\theta}}^{\pi}\gets\tau\bm{\theta}^{\pi}+(1-\tau)\bar{\bm{\theta}}^{\pi}$
\ENDIF
\ENDWHILE
\end{algorithmic}
\end{algorithm}

\section{Examples of Prompts and Responses for GPT-4}
\label{SA}
\subsection{An Example of Prompt Design With Human Feedback}
The following dialogues show the prompts and responses of OpenAI GPT-4 for the task \texttt{PandaPickAndPlace} in the \texttt{panda\_gym} environment. \\

\img{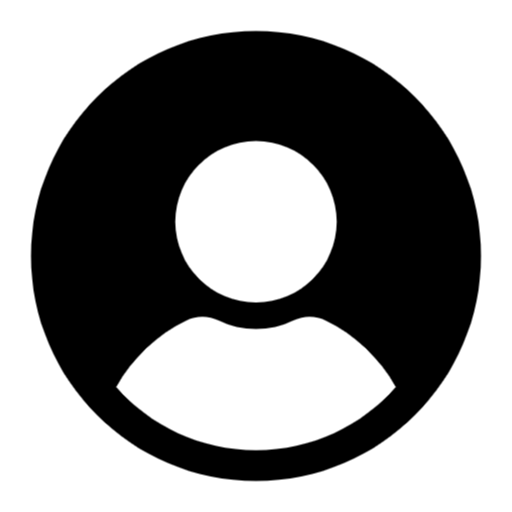} User
\vspace{-0.1cm}
\begin{center}
\begin{tcolorbox}[breakable, enhanced, width=0.96\textwidth, opacityfill=1., colback=white]
Think about you are an expert who would like to finish a task with the Franka Emika Panda Robot. Please provide a step-by-step description of how the end effector should move in order to finish this task.\\

[Start of the General Task Description]

This task is to use the end effector of the Franka Emika Panda Robot to pick one object from one assigned location on the table and place this object to another location. The object to be moved is a cube with a side of 0.04 m. 

[End of the General Task Description]\\

[Start of the Question]

How many phases can the end effector movement be divided into? 

[End of the Question]\\

[Start of the Template]

Phase [NUM]: The gripper should move [CHOICE: to, around, above] [LOC].\\
Phase [NUM]: The gripper should be [CHOICE: open, closed].

[End of the Template]\\

Rules:\\
1. Please use the above template to answer, but do not include “[Start of the Template]” and “[End of the Template]” in your response. \\
2. If you see phrases like [NUM], replace the entire phrase with an integer. \\
3. If you see phrases like [LOC], replace the entire phrase with a location.\\
4. If you see phrases like [CHOICE: choice1, choice2, ...], you should replace the entire phrase with one of the choices listed.\\
5. If you see [OPTIONAL], it means you only add the content if necessary.\\
6. You do not need to ensure that the object is moved to the desired position successfully.
\end{tcolorbox}
\end{center}

\hfill GPT-4 \img{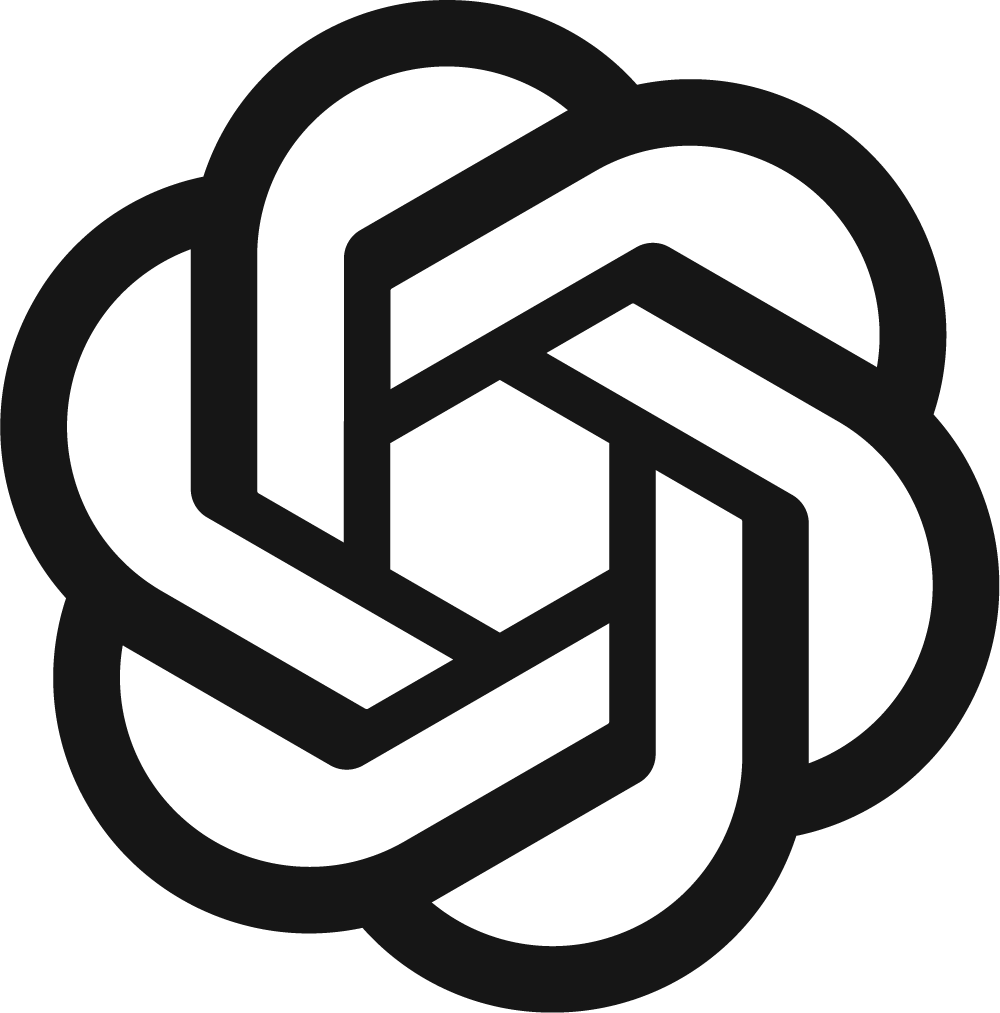}~~
\vspace{-0.1cm}
\begin{center}
\begin{tcolorbox}[breakable, enhanced, width=0.96\textwidth, opacityfill=1., colback=white]
Phase 1: The gripper should move above the assigned location on the table where the object is placed.\\
Phase 2: The gripper should be open.\\
Phase 3: The gripper should move to the object.\\
Phase 4: The gripper should be closed around the object to ensure a secure grip.\\
Phase 5: The gripper should move above the initial location to clear any possible obstructions on the table.\\
Phase 6: The gripper should move above the new designated location on the table where the object is to be placed.\\
Phase 7: The gripper should move to the new location on the table.\\
Phase 8: The gripper should be open to release the object.\\
Phase 9: The gripper should move above the new location to clear the object and complete the task.
\end{tcolorbox}
\end{center}

\img{figures/user_logo.png} User
\vspace{-0.1cm}
\begin{center}
\begin{tcolorbox}[breakable, enhanced, width=0.96\textwidth, opacityfill=1., colback=white]
Can you write the detailed codes for the controller of this process? You can write the staged controller in a nested if-else statement if necessary. \\

The controller inputs are the following variables.\\
1. gripper\_position (dim: 3)\\
2. gripper\_speed (dim:3)\\
3. gripper\_finger\_distance (dim: 1)\\
4. object\_position (dim: 3)\\
5. object\_speed (dim: 3)\\
6. object\_orientation (dim: 3)\\
7. object\_rotational\_speed (dim: 3)\\
8. target\_position (dim: 3)\\

You must consider all the following information to make the task less ambiguous: \\
1. The output of the controller should be a 4-dimensional vector. The first three elements of this vector are the\\ 3-dimensional gripper displacement and the last element is the distance changing value between the fingers.\\
2. The maximum moving distance of the gripper in each dimension is 0.05 m at one step. \\
3. When the finger distance is smaller than 0.04 m, the gripper can be thought of as closed. When the finger is larger than 0.06 m, the gripper can be thought of as open. The finger distance ranges from 0 to 0.1 m. You can try to use the largest allowable finger movement when changing its states in order to guarantee the finger state is indeed changed. \\
4. If you would like to check if the gripper is proximal to the object, you can use the proximal distance of 0.02 m. \\
5. The initial state of the gripper might be either closed or open. You should check the status and open it if necessary. \\
6. You do not need to clearly indicate the phases, but you may use them implicitly to write the controller codes. \\
7. You may use numpy when necessary. \\
8. You only need to write the detailed controller codes and do not need to provide examples. 
\end{tcolorbox}
\end{center}

\newpage
\hfill GPT-4 \img{figures/gpt_logo.png}~~
\vspace{-0.1cm}
\begin{center}
\begin{tcolorbox}[breakable, enhanced, width=0.96\textwidth, opacityfill=1., colback=white]
Given the complexities involved in robotic control, it's ideal to use a state machine for managing the actions of the robot over time, however, the task asks for a nested if-else approach which can be less clear and more prone to errors. Below is a simplistic approach using if-else control structure. It's recommended to further refine and test this code in a simulated environment before attempting to use it with a real robot.
\begin{lstlisting}[language=Python]
import numpy as np

def controller(gripper_position, gripper_speed, gripper_finger_distance, object_position, object_speed, object_orientation, object_rotational_speed, target_position):
    # Constants
    MAX_MOVE_DISTANCE = 0.05
    PROXIMAL_DISTANCE = 0.02
    CLOSE_FINGER_DISTANCE = 0.04
    OPEN_FINGER_DISTANCE = 0.06
    FINGER_MAX_MOVE = 0.05
    
    # Output: [dx, dy, dz, dfinger]
    control_output = np.zeros(4)
    
    # Calculate the euclidean distances
    distance_to_object = np.linalg.norm(gripper_position - object_position)
    distance_to_target = np.linalg.norm(gripper_position - target_position)
    
    # Check the initial state of the gripper and open if necessary
    if gripper_finger_distance < OPEN_FINGER_DISTANCE:
        control_output[3] = FINGER_MAX_MOVE  # Open the gripper to the max allowable finger movement
    else:
        # If the gripper is above or near the object
        if distance_to_object <= PROXIMAL_DISTANCE:
            # If the gripper is open, close it to pick the object
            if gripper_finger_distance > CLOSE_FINGER_DISTANCE:
                control_output[3] = -FINGER_MAX_MOVE  # Close the gripper to the max allowable finger movement
            else:
                # If gripper is closed, move towards the target
                direction_to_target = (target_position - gripper_position) / distance_to_target
                move_distance = min(distance_to_target, MAX_MOVE_DISTANCE)
                control_output[:3] = direction_to_target * move_distance
        else:
            # If the gripper is not near the object, move towards the object
            direction_to_object = (object_position - gripper_position) / distance_to_object
            move_distance = min(distance_to_object, MAX_MOVE_DISTANCE)
            control_output[:3] = direction_to_object * move_distance
    
    # Ensure the finger movement is within the range [0, 0.1]
    control_output[3] = np.clip(control_output[3] + gripper_finger_distance, 0, 0.1) - gripper_finger_distance
    
    return control_output
\end{lstlisting}
In this code, we have assumed that the robot should move in a straight line towards the object and then towards the target. The `controller' function calculates the control output based on the current state of the robot and the object. The control output is a 4D vector where the first three elements are the displacement of the gripper and the last element is the change in finger distance. This code does not handle obstacles, complex trajectories, or unexpected behaviors of the robot or the object. Moreover, this code does not handle the orientation of the gripper or the object, and assumes the object will be in the correct orientation for grasping and placing.
\end{tcolorbox}
\end{center}

\img{figures/user_logo.png} User
\vspace{-0.1cm}
\begin{center}
\begin{tcolorbox}[breakable, enhanced, width=0.96\textwidth, opacityfill=1., colback=white]
After grasping an object, you should hold it firmly and move it to the target position. Can you fine-tune the codes above by incorporating this information? 
\end{tcolorbox}
\end{center}

\newpage
\hfill GPT-4 \img{figures/gpt_logo.png}~~
\vspace{-0.1cm}
\begin{center}
\begin{tcolorbox}[breakable, enhanced, width=0.96\textwidth, opacityfill=1., colback=white]
Certainly! In order to ensure that the gripper holds the object firmly while moving it to the target position, it is important to maintain the gripper's closed state once the object has been grasped. In the updated code below, I added a boolean variable `object\_grasped' to keep track of whether the object has been grasped or not. Once the object is grasped, the `object\_grasped' variable is set to `True', and the gripper will maintain its closed state as it moves to the target position.
\begin{lstlisting}[language=Python]
import numpy as np

def controller(gripper_position, gripper_speed, gripper_finger_distance, object_position, object_speed, object_orientation, object_rotational_speed, target_position):
    # Constants
    MAX_MOVE_DISTANCE = 0.05
    PROXIMAL_DISTANCE = 0.02
    CLOSE_FINGER_DISTANCE = 0.04
    OPEN_FINGER_DISTANCE = 0.06
    FINGER_MAX_MOVE = 0.05
    
    # Output: [dx, dy, dz, dfinger]
    control_output = np.zeros(4)
    
    # Calculate the euclidean distances
    distance_to_object = np.linalg.norm(gripper_position - object_position)
    distance_to_target = np.linalg.norm(gripper_position - target_position)
    
    # Boolean variable to check if the object has been grasped
    object_grasped = gripper_finger_distance <= CLOSE_FINGER_DISTANCE and distance_to_object <= PROXIMAL_DISTANCE
    
    # Check the initial state of the gripper and open if necessary
    if gripper_finger_distance < OPEN_FINGER_DISTANCE and not object_grasped:
        control_output[3] = FINGER_MAX_MOVE  # Open the gripper to the max allowable finger movement
    else:
        # If the gripper is above or near the object
        if distance_to_object <= PROXIMAL_DISTANCE and not object_grasped:
            # If the gripper is open, close it to pick the object
            if gripper_finger_distance > CLOSE_FINGER_DISTANCE:
                control_output[3] = -FINGER_MAX_MOVE  # Close the gripper to the max allowable finger movement
            else:
                object_grasped = True  # Set object_grasped to True as the gripper is now closed and near the object
        elif object_grasped:
            # If object is grasped, move towards the target
            direction_to_target = (target_position - gripper_position) / distance_to_target
            move_distance = min(distance_to_target, MAX_MOVE_DISTANCE)
            control_output[:3] = direction_to_target * move_distance
        else:
            # If the gripper is not near the object, move towards the object
            direction_to_object = (object_position - gripper_position) / distance_to_object
            move_distance = min(distance_to_object, MAX_MOVE_DISTANCE)
            control_output[:3] = direction_to_object * move_distance
    
    # Ensure the finger movement is within the range [0, 0.1]
    control_output[3] = np.clip(control_output[3] + gripper_finger_distance, 0, 0.1) - gripper_finger_distance
    
    return control_output
\end{lstlisting}
In this updated code, the `object\_grasped' variable is checked in the if-else control structure to decide the next actions. Once the object is grasped, the `object\_grasped' variable ensures that the gripper remains closed and moves towards the target position.
\end{tcolorbox}
\end{center}

\newpage
\subsection{An Example of Prompt Design With A Code Template}
The following dialogues show the prompts and responses of OpenAI GPT-4 for the task \texttt{meat\_off\_grill} in the \texttt{RLBench} environment. We use the code for the task \texttt{push\_button} as a template and three pre-defined reusable primitive functions \texttt{is\_close}, \texttt{normalize\_euler\_angle}, and \texttt{get\_action} are called by the template.\\

\img{figures/user_logo.png} User
\vspace{-0.1cm}
\begin{center}
\begin{tcolorbox}[breakable, enhanced, width=0.96\textwidth, opacityfill=1., colback=white]
Think about you are an expert who would like to finish a task with the Franka Emika Panda Robot. The controlled variables are a 7-dimensional variable of the gripper, which consists of a 3-dimensional gripper displacement, a 3-dimensional gripper orientation change value described by the Euler angle, and a 1-dimensional element indicating if the gripper is open or closed. Please provide a step-by-step description of how the end effector should move in order to finish this task. \\

[Start of the General Task Description] 

This task is to use the end effector of the Franka Emika Panda Robot to grasp a chicken from a grill and then put to a specified location. 

[End of the General Task Description]\\

[Start of the Question]

How many phases the end effector movement can be divided into? 

[End of the Question]\\

[Start of the Template]

Phase [NUM]: The gripper should move [CHOICE: to, around] [LOC] [OPTIONAL: and [CHOICE: open, close] its gripper]. 

[End of the Template]\\

Rules:\\
1. Please use the above template to answer, but do not include “[Start of the Template]” and “[End of the Template]” in your response. \\
2. If you see phrases like [NUM], replace the entire phrase with an integer. \\
3. If you see phrases like [LOC], replace the entire phrase with a location.\\
4. If you see phrases like [CHOICE: choice1, choice2, ...], you should replace the entire phrase with one of the choices listed.\\
5. If you see [OPTIONAL], it means you only add the content if necessary.\\
6. You do not need to ensure that the object is moved to the desired position successfully. \\
7. You should not add redundant phases in your responses. 
\end{tcolorbox}
\end{center}

\hfill GPT-4 \img{figures/gpt_logo.png}~~
\vspace{-0.1cm}
\begin{center}
\begin{tcolorbox}[breakable, enhanced, width=0.96\textwidth, opacityfill=1., colback=white]
The end effector movement for the task described can be divided into the following phases:\\
Phase 1: The gripper should move to above the chicken and open its gripper.\\
Phase 2: The gripper should move to the chicken.\\
Phase 3: The gripper should close its gripper.\\
Phase 4: The gripper should move to above the grill with the chicken.\\
Phase 5: The gripper should move to the specified location.\\
Phase 6: The gripper should open its gripper.\\
Phase 7: The gripper should move away from the specified location.
\end{tcolorbox}
\end{center}

\img{figures/user_logo.png} User
\vspace{-0.1cm}
\begin{center}
\begin{tcolorbox}[breakable, enhanced, width=0.96\textwidth, opacityfill=1., colback=white]
Think about you are a Python programming expert. Can you write the detailed codes for the controller of this process? \\

The controller inputs include the following variables. These variables are all np.ndarray with shape (6,). The first 3 elements indicate the locations and the last 3 elements indicate Euler angles expressing the orientation. \\

1. ‘gripper\_state’: The current gripper state\\
2. ‘chicken\_state’: The current chicken state\\
3. ‘target\_chicken\_position’: The desired position the chicken should be put to\\
4. ‘gripper\_open’: Indicator if the gripper is open or not\\

Note that you can use only these 3 variables and should not assume any other additional variances. You should use the following code template for push\_button task to generate the codes for this task. For the push\_button task, the goal is to push a button with the gripper. \\

\#\#\# Start of The Code Template \#\#\#
\begin{lstlisting}[language=Python]
def get_action(button_state, gripper_state, position_above_button, action_limit=0.05):
    # For the task of push_button, we do not care the rotation around the z-axis since this rotation will not influence the push effects.
    # Thus, in this case, we set a euler_mask to mask out the possible rotations around the z-axis.
    euler_mask = np.array([1., 1., 0.])

    # There are two phases for the push_button task, i.e., 1) putting the gripper above the button and 2) pressing the gripper
    # The first step for the controller is to identify the current stages. In order to do this, we create two variables 
    # "gripper_above_button" to indicate where the gripper is above the button and "ready_to_press" to indicate whether the gripper
    # is ready to push the button. These variables are initialized to be False's, and the values of them will be determined by 
    # following conditions. 
    gripper_above_button = False
    ready_to_press = False

    # is_close is function that decides if a three-dimensional np.ndarray is close to another three-dimensional np.ndarray. These
    # np.ndarray's can be either a position and a set of Euler angles.
    # You should directly use the function "normalize_euler_angle" to normalize the Euler angles since some angles may not be within 
    # appropriate ranges. 
    # Be sure to add the gripper_open condition in the conditions below. This may be necessary for some other tasks. 
    if is_close(gripper_state[:3], position_above_button[:3]) \
        and is_close(normalize_euler_angle(gripper_state[3:] - position_above_button[3:]) * euler_mask, np.zeros(3)):
        gripper_above_button = True

    if is_close(gripper_state[:3], button_state[:3]) \
        and is_close(normalize_euler_angle(gripper_state[3:] - button_state[3:]) * euler_mask, np.zeros(3)) and gripper_open is False:
            # In order for ready_to_press to be True, we should have two prerequisite: 
            # 1) The gripper should be close to the button. 
            # 2) gripper_open should be True.
            ready_to_press = True

    # Based on the phase conditions, we use get_action function to generate the next actions. You do not need to care the meanings of
    # action and buffer_action, which are the outputs of the function get_action. The inputs of the function get_action include:
    # 1) cur_gripper_pose: The current gripper pose. Default to gripper_state.
    # 2) target_gripper_pose: The target gripper pose in this case, default to be target. Note that the target should be carefully
    # assigned for different phases.
    # 3) gripper_open: Indicator where the gripper should be open or not in the next action. You should decide if it should be True or False
    # 4) action_limit: The maximum value of the displacement and Euler angles' changing values per dimension. Default to action_limit.
    # 5) euler_mask: The mask of Euler angles' changes. Default to euler_mask. 
    # NOTE: You should pay attention to the order of the conditions below. A wrong order will completely fail the task. 
    # NOTE: You should use all the phase indicators above. For example, you should use both "ready_to_press" and "gripper_above_button" 
    # as conditions
    if ready_to_press:
        # If ready_to_press is True, then we do not need to care if gripper_above_button is True or False. In this case, we can directly
        # push the button by setting target_gripper_pose to button_state.
        target = position_below_button
        action, buffer_action = get_action(cur_gripper_pose=gripper_state, 
                                           target_gripper_pose=target, gripper_open=False, 
                                           action_limit=action_limit, euler_mask=euler_mask)
    elif gripper_above_button:
        # If ready_to_press is False but gripper_above_button is True, then we should close the gripper since we should use a closed 
        # gripper to push the button.
        target = button_state
        action, buffer_action = get_action(cur_gripper_pose=gripper_state, 
                                           target_gripper_pose=target, gripper_open=False, 
                                           action_limit=action_limit, euler_mask=euler_mask)
    else:
        # If both ready_to_press and gripper_above_button are False's, then we need to move the gripper to position_above_button. 
        target = position_above_button
        action, buffer_action = get_action(cur_gripper_pose=gripper_state, 
                                           target_gripper_pose=target, gripper_open=True, 
                                           action_limit=action_limit, euler_mask=euler_mask)

    return action, buffer_action
\end{lstlisting}
\#\#\# End of The Code Template \#\#\#\\\\

Here are some information you must consider:\\
1. When the gripper is ready to grasp an object, the gripper must be open and the gripper position should be the same as the object position. \\
2. For the task meat\_off\_grill, you do not need to care the roration of Euler angles around the x-axis. That is, you can let euler\_mask to be np.array([0., 1., 1.]). \\
3. You can assume that the functions is\_close, normalize\_euler\_angle, and get\_action are available. In order words, you do not need to write the details of these functions. 
\end{tcolorbox}
\end{center}

\hfill GPT-4 \img{figures/gpt_logo.png}~~
\vspace{-0.1cm}
\begin{center}
\begin{tcolorbox}[breakable, enhanced, width=0.96\textwidth, opacityfill=1., colback=white]
Certainly! Following the code template and guidelines provided, here's how you might write the controller code for the task of moving a chicken off the grill with the Franka Emika Panda Robot. This script will assume that `is\_close', `normalize\_euler\_angle', and `get\_action' are predefined functions available in the context.
\begin{lstlisting}[language=Python]
import numpy as np

def get_action_for_chicken(chicken_state, gripper_state, target_chicken_position, action_limit=0.05):
    euler_mask = np.array([0., 1., 1.])  # Mask out rotation around the x-axis

    # Phase indicators
    gripper_above_chicken = False
    gripper_ready_to_grasp = False
    gripper_holding_chicken = False
    chicken_at_target = False

    # Check if the gripper is above the chicken
    if is_close(gripper_state[:3], chicken_state[:3]) and gripper_open:
        gripper_above_chicken = True
    
    # Check if the gripper is in the right position to grasp the chicken
    if gripper_above_chicken and is_close(normalize_euler_angle(gripper_state[3:] - chicken_state[3:]) * euler_mask, np.zeros(3)):
        gripper_ready_to_grasp = True

    # Check if the gripper is holding the chicken (gripper is not open and it's close to the chicken state)
    if not gripper_open and is_close(gripper_state[:3], chicken_state[:3]):
        gripper_holding_chicken = True

    # Check if the chicken has been placed at the target position
    if gripper_holding_chicken and is_close(gripper_state[:3], target_chicken_position[:3]):
        chicken_at_target = True

    # Decision making based on phase indicators
    if chicken_at_target:
        # Open the gripper to release the chicken
        action, buffer_action = get_action(cur_gripper_pose=gripper_state, 
                                           target_gripper_pose=target_chicken_position, 
                                           gripper_open=True, 
                                           action_limit=action_limit, 
                                           euler_mask=euler_mask)
    elif gripper_holding_chicken:
        # Move to the target position with the chicken
        action, buffer_action = get_action(cur_gripper_pose=gripper_state, 
                                           target_gripper_pose=target_chicken_position, 
                                           gripper_open=False, 
                                           action_limit=action_limit, 
                                           euler_mask=euler_mask)
    elif gripper_ready_to_grasp:
        # Close the gripper to grasp the chicken
        action, buffer_action = get_action(cur_gripper_pose=gripper_state, 
                                           target_gripper_pose=chicken_state, 
                                           gripper_open=False, 
                                           action_limit=action_limit, 
                                           euler_mask=euler_mask)
    elif gripper_above_chicken:
        # Descend to the chicken position to be ready to grasp
        action, buffer_action = get_action(cur_gripper_pose=gripper_state, 
                                           target_gripper_pose=chicken_state, 
                                           gripper_open=True, 
                                           action_limit=action_limit, 
                                           euler_mask=euler_mask)
    else:
        # Move above the chicken
        action, buffer_action = get_action(cur_gripper_pose=gripper_state, 
                                           target_gripper_pose=np.concatenate((chicken_state[:3], gripper_state[3:])), 
                                           gripper_open=True, 
                                           action_limit=action_limit, 
                                           euler_mask=euler_mask)

    return action, buffer_action
\end{lstlisting}
This code defines a function that decides the next action the robot should take based on the current state of the gripper and chicken and the target position for the chicken. The function updates several boolean flags based on the current situation, then makes a decision on the next action to take. It uses the predefined `is\_close', `normalize\_euler\_angle', and `get\_action' functions to calculate whether the gripper is in the correct position relative to the chicken and target and to generate the appropriate action commands.
\end{tcolorbox}
\end{center}

The code of the primitive \texttt{get\_action}, which is called by \texttt{get\_action\_for\_chicken}, can also be generated correctly by GPT-4. The details are given as follows. \\

\img{figures/user_logo.png} User
\vspace{-0.1cm}
\begin{center}
\begin{tcolorbox}[breakable, enhanced, width=0.96\textwidth, opacityfill=1., colback=white]
Think about you are an expert who would like to finish a task with the Franka Emika Panda Robot. The controlled variables are a 7-dimensional variable of the gripper, which consists of a 3-dimensional gripper displacement, a 3-dimensional gripper orientation change value described by the Euler angle, and a 1-dimensional element indicating if the gripper is open or closed. \\

Please write a primitive function with the name “get\_action” that decides the gripper action based on the current gripper pose and target gripper pose. You should use the Python language. \\

This primitive function should have the following inputs. \\
1. cur\_gripper\_pose: (np.ndarray of shape (6,)) \\
The current gripper pose with the first 3 elements being the position and the last 3 elements being the Euler angles describing the orientation \\
2. target\_gripper\_pose: (np.ndarray of shape (6,)) \\
The target gripper pose with the first 3 elements being the target position and the last 3 elements being the Euler angles describing the target orientation \\
3. gripper\_open: (bool) \\
The variable describing whether the gripper should be open or not \\
4. action\_limit: (float, default to 0.05) \\
The maximum gripper displacement in each of the 3 dimensions. Note that you can multiply the this limit with ‘multiplier\_for\_euler\_angle\_limit’ and use the product to clip the maximum changing values of Euler angles \\
5. symmetric\_z: (bool, default to True) \\
The indicator of whether the rotation around the z-axis can be regarded symmetric or not when using the normalize\_euler\_angle function \\
6. euler\_mask (np.ndarray of shape (3,), default to np.ones(3)) \\
For some tasks the rotation around some specific axes is not necessary. In this case, we can use a mask to mask out the unnecessary rotations.\\

The outputs of this function include the following two variables. \\
1. A np.ndarray of shape (8,), which should be named “action”. \\
2. A np.ndarray of shape (7,), which should be named “buffer\_action”. \\

For the first output with the name “action”, we have the following requirements. \\
1. The first three elements action[:3] should describe the next gripper position subject to the dimension-wise action limit with respect to the displacement. \\
2. The next four elements action[3:7] should describe the next gripper quaternion subject to the dimension-wise action limit with respect to the Euler angle changes. Note that this quaternion should correspond to the absolute Euler angles after changing rather than the incremental Euler angles. \\
3. The last element action[-1] describes whether the gripper should be open or not. If the gripper should be open, then this element should be 1, else it should be 0.\\

For the second output with the name “buffer\_action”, we have the following requirements. \\
1. The first three elements buffer\_action[:3] should describe the gripper displacement subject to the dimension-wise action limit. However, it should be normalized to the range [-1, 1]. \\
2. The next three elements buffer\_action[3:6] should describe the Euler angle changes subject to the dimension-wise action limit. However, it should be normalized to the range [-1, 1]. \\
3. The last element buffer\_action[-1] describes whether the gripper should be open or not. If the gripper should be open, then this element should be 1, else it should be -1. \\

NOTE: \\
1. You can directly use the function “euler\_to\_quaternion” to convert the 3-dimensional Euler angles to a 4-dimensional quaternion. You do not need to show this function in your response. \\
2. You may directly use the function “normalize\_euler\_angle” to normalize the Euler angles since some angles may not be within appropriate ranges. For example, some Euler angles may be larger than 2*pi rad. You should implement “normalize\_euler\_angle” before the dimension-wise action limit. You do not need to show this function in your response. 
\end{tcolorbox}
\end{center}

\hfill GPT-4 \img{figures/gpt_logo.png}~~
\vspace{-0.1cm}
\begin{center}
\begin{tcolorbox}[breakable, enhanced, width=0.96\textwidth, opacityfill=1., colback=white]
To create the `get\_action' function as described, we'll need to consider the following steps:\\

\begin{itemize}
\item [1.] \textbf{Calculate Gripper Displacement and Euler Angle Changes:} Determine the difference between the current and target poses for both position and Euler angles.

\item [2.] \textbf{Normalize Euler Angles:} Use the `normalize\_euler\_angle' function to ensure the Euler angles are within appropriate ranges, considering the `symmetric\_z' flag and `euler\_mask'.

\item [3.] \textbf{Apply Dimension-wise Action Limits:} Clip the calculated displacements and Euler angle changes within the specified limits.

\item [4.] \textbf{Convert Euler Angles to Quaternion:} Use the `euler\_to\_quaternion' function to convert the adjusted Euler angles to a quaternion.

\item [5.] \textbf{Normalize for Buffer Action:} Normalize the displacement and Euler angle changes to the range [-1, 1].

\item [6.] \textbf{Assemble the `action' and `buffer\_action' Arrays:} Construct these arrays based on the calculated values and the `gripper\_open' flag.
\end{itemize}
~\\
Here's the implementation of the `get\_action' function:
\begin{lstlisting}[language=Python]
import numpy as np

def get_action(cur_gripper_pose, target_gripper_pose, gripper_open, action_limit=0.05, 
               symmetric_z=True, euler_mask=np.ones(3)):
    # Calculate displacement and Euler angle changes
    displacement = target_gripper_pose[:3] - cur_gripper_pose[:3]
    euler_change = target_gripper_pose[3:] - cur_gripper_pose[3:]

    # Normalize Euler angles
    normalized_euler = normalize_euler_angle(euler_change, symmetric_z)
    normalized_euler *= euler_mask

    # Apply action limits
    limited_displacement = np.clip(displacement, -action_limit, action_limit)
    euler_angle_limit = action_limit * multiplier_for_euler_angle_limit
    limited_euler_change = np.clip(normalized_euler, -euler_angle_limit, euler_angle_limit)

    # Convert normalized and limited Euler angles to quaternion
    next_euler_angles = cur_gripper_pose[3:] + limited_euler_change
    quaternion = euler_to_quaternion(next_euler_angles)

    # Normalize for buffer action
    norm_displacement = limited_displacement / action_limit
    norm_euler_change = limited_euler_change / euler_angle_limit

    # Assemble action array
    action = np.zeros(8)
    action[:3] = cur_gripper_pose[:3] + limited_displacement
    action[3:7] = quaternion
    action[-1] = 1 if gripper_open else 0

    # Assemble buffer action array
    buffer_action = np.zeros(7)
    buffer_action[:3] = norm_displacement
    buffer_action[3:6] = norm_euler_change
    buffer_action[-1] = 1 if gripper_open else -1

    return action, buffer_action
\end{lstlisting}
Note:
\begin{itemize}
\item The `euler\_to\_quaternion' and `normalize\_euler\_angle' functions are assumed to be defined elsewhere as mentioned in your requirements.
\item `multiplier\_for\_euler\_angle\_limit' should be defined or passed as a parameter if it's variable. If it's a constant, you can directly use its value in the code.
\end{itemize}
\end{tcolorbox}
\end{center}

\color{black}
\section{Applicability Analysis of RLingua and Simulation Task Selections and Descriptions}
\label{SC}
This appendix section provides an analysis of the practical applicability of the proposed algorithm, RLingua, detailing scenarios where it is effective and where it may not be suitable. Additionally, The criteria used for selecting tasks in \texttt{RLBench} are also illustrated.
\subsection{Applicability Analysis of RLingua}
\label{SCA}
RLingua leverages the prior knowledge of LLMs and RL to obtain robot controllers with high performance. The success of the robot controller depends on the following two aspects.

\begin{itemize}
\item [1)] The capacity of the current LLM (specifically, GPT-4 in our study) to generate a rule-based controller endowed with accurate control logic.
\item [2)] The robot task should not be brittle to RL explorations. 
\end{itemize}

With the task examples in \texttt{RLBench}, we illustrate these two aspects in detail as follows. 
\begin{itemize}
\item [1)] RLingua may fail some tasks due to the currently limited ability of GPT-4. Specifically, GPT-4's failure to generate a rule-based controller with correct control logic can be attributed primarily to the following reasons.
\begin{itemize}
\item [i)] GPT-4 is currently unable to generate gripper paths that include arcs. Instead, it is limited to producing paths with polyline shapes, lacking the capability to handle the more complex calculations required for arcs. This limitation becomes apparent in certain tasks within \texttt{RLBench}, such as \texttt{close\_door} and \texttt{close\_fridge}, where an arc segment in the gripper's trajectory is necessary. Consequently, for these tasks, GPT-4 fails to generate a rule-based controller capable of providing reliable guidance for RL. 

\item [ii)] GPT-4 struggles with robot tasks that feature long horizons. The Python-written rule-based controllers it generates rely heavily on multiple, and often nested, \texttt{if-else} statements to recognize the current task state and determine subsequent control actions. When a task encompasses an extensive horizon with numerous stages, GPT-4's capability to produce a controller with accurate logic is compromised, as it requires the correctness of \textit{all} \texttt{if-else} statements. Tasks from \texttt{RLBench} falling into this category include, but are not limited to, \texttt{block\_pyramid}, \texttt{stack\_blocks}, and \texttt{wipe\_desk}.

\item [iii)] The controller generated by GPT-4 may sometimes fail to identify the correct grasp pose for specific tasks. In tasks within \texttt{RLBench}, such as \texttt{tv\_on} and \texttt{change\_channel}, the robot gripper must accurately find a correct grasp pose. During our experiments, we observed instances where the GPT-4 generated controller was unable to reliably grasp an object in the correct pose. For instance, given the symmetric nature of the gripper, grasping an object at joint angles of $90^{\circ}$ and $-90^{\circ}$ should be equivalent. However, the current GPT-4 model struggles to effectively integrate this crucial piece of information into the controller code generation process. This oversight can lead to violations of the joint rotation constraints for the Franka Emika Panda robot, resulting in task failure. Consequently, these failures indicate that GPT-4 is unable to generate a controller with the logical reliability necessary to guide reinforcement learning exploration effectively. 
\end{itemize}

\item [2)] RLingua is not ideally suited for tasks that are particularly vulnerable to the dynamics of RL exploration. 

Effective exploration is crucial for achieving optimal performance in RL. Nonetheless, certain tasks within \texttt{RLBench} exhibit a high sensitivity to these exploration processes. Take, for instance, the \texttt{stack\_cups} task, which demands that the robot sequentially pick two cups and stack them into a third one. For successful execution, the gripper must grasp each cup by its brim. However, during RL exploration phases, a cup may be inadvertently released at any point post-grasp, leading to a high likelihood of the cup tipping over. Subsequent attempts to re-grasp a tipped cup significantly compound the challenge, rendering the task exceedingly difficult due to the sparsity of successful outcomes.
\end{itemize}

In summary, the effectiveness of RLingua hinges on two primary factors: the capacity of current LLMs to generate rule-based robot controllers, and the susceptibility of tasks to the challenges posed by RL exploration. Regarding the first factor, there is optimism that LLM capabilities will see substantial advancements shortly, thereby extending RLingua's applicability to more complex tasks. As for the task's brittleness to RL explorations, leveraging LLMs' prior knowledge to minimize unnecessary explorations and circumvent actions leading to unrecoverable states presents a viable strategy. An illustrative example is the known prerequisite in the \texttt{stack\_cups} task within \texttt{RLBench}, where a cup should not be released post-grasp until it is positioned above the target. By integrating such knowledge, we can programmatically restrict the gripper from releasing the cup during transit, thereby eliminating certain exploration actions. Looking ahead, our future work aims to refine the prompt design for LLMs, enabling the extraction of such task-specific knowledge directly from LLMs rather than relying on human input. This ambition points towards a deeper integration of LLMs with RL, a direction that promises exciting developments and is earmarked for subsequent investigation.

\subsection{Simulation Task Selections and Descriptions}
\label{SCB}
This section outlines the criteria used for selecting tasks in our simulation experiments and provides detailed introductions to the specific robot tasks in our experiments.
\subsubsection{\texttt{RLBench} Simulation Task Selections and Descriptions}
Based on the applicability analysis in Section \ref{SCA}, we classify the 108 robot tasks in \texttt{RLBench} \cite{james2020rlbench} into different categories, as shown in TABLE \ref{TA2}. There are 34 tasks that are considered for RLingua and we randomly sample 12 tasks out of them in the experiment in Section \ref{S42}. To guarantee task diversity, we consider the following criteria when sampling the tasks. 

\begin{itemize}
\item [--] Reach a target (1 task): \texttt{push\_button}
\item [--] Pick and lift (2 tasks): \texttt{take\_lid\_off\_saucepan}, \texttt{take\_umbrella\_out\_of\_umbrella\_stand}
\item [--] The task requiring dynamic interactions (1 task): \texttt{slide\_block\_to\_target}
\item [--] Pick and place (4 tasks): \texttt{close\_jar}, \texttt{meat\_off\_grill}, \texttt{phone\_on\_base}, \texttt{put\_item\_in\_container}
\item [--] The tasks requiring complex gripper rotations (4 tasks): \texttt{close\_drawer}, \texttt{open\_drawer}, \texttt{take\_usb\_out\_of\_com-\\puter}, \texttt{unplug\_charger}
\end{itemize}

\begin{table}[!t]
\centering
\captionsetup{justification=centering}
\tabletag{A-I}
\caption{\color{black}The classification of robot tasks in \texttt{RLBench}}
\label{TA2}
\color{black}
\begin{tabular}{wc{2cm}wc{15cm}}
\hline
Category & Tasks \\
\hline
\makecell{Tasks considered\\ by RLingua} &
\makecell{\texttt{basketball\_in\_hoop}, 
\textbf{\texttt{close\_drawer}},
\textbf{\texttt{close\_jar}},
\texttt{close\_microwave},
\texttt{insert\_onto\_square\_peg},
\texttt{lamp\_off},\\
\texttt{lamp\_on},
\texttt{lift\_numbered\_block},
\texttt{light\_bulb\_in},
\texttt{light\_bulb\_out},
\textbf{\texttt{meat\_off\_grill}},
\texttt{meat\_on\_grill},\\
\textbf{\texttt{open\_drawer}},
\texttt{open\_jar},
\texttt{open\_wine\_bottle},
\texttt{pick\_and\_lift},
\texttt{pick\_and\_lift\_small},
\texttt{pick\_up\_cup},\\
\textbf{\texttt{phone\_on\_base}},
\texttt{place\_shape\_in\_shape\_sorter},
\texttt{play\_jenga},
\texttt{press\_switch},
\textbf{\texttt{push\_button}},
\texttt{push\_buttons},\\
\textbf{\texttt{put\_item\_in\_container}},
\texttt{reach\_target},
\textbf{\texttt{slide\_block\_to\_target}},
\texttt{solve\_puzzle},
\textbf{\texttt{take\_lid\_off\_saucepan}},\\
\texttt{take\_off\_weighing\_scales},
\texttt{take\_toilet\_roll\_off\_stand},
\textbf{\texttt{take\_umbrella\_out\_of\_umbrella\_stand}},\\
\textbf{\texttt{take\_usb\_out\_of\_computer}},
\textbf{\texttt{unplug\_charger}}}  \\
\hline
\makecell{An arc in the\\ gripper path} & 
\makecell{\texttt{close\_door},
\texttt{close\_fridge},
\texttt{close\_grill},
\texttt{close\_laptop\_lid},
\texttt{close\_box},
\texttt{empty\_dishwasher},
\texttt{open\_box},\\
\texttt{open\_door},
\texttt{open\_fridge},
\texttt{open\_grill},
\texttt{open\_microwave}, 
\texttt{open\_oven},
\texttt{open\_washing\_machine},
\texttt{open\_window},\\
\texttt{put\_shoes\_in\_box},
\texttt{put\_tray\_in\_oven},
\texttt{take\_shoes\_out\_of\_box}, 
\texttt{take\_tray\_out\_of\_oven},\\ 
\texttt{toilet\_seat\_down},
\texttt{toilet\_seat\_up},
\texttt{turn\_oven\_on}} \\
\hline
\makecell{Too long task \\horizon for LLMs} & 
\makecell{\texttt{block\_pyramid},
\texttt{empty\_container},
\texttt{empty\_dishwasher},
\texttt{get\_ice\_from\_fridge},
\texttt{open\_window},\\
\texttt{put\_item\_in\_drawer},
\texttt{stack\_blocks},
\texttt{straighten\_rope},
\texttt{take\_item\_out\_of\_drawer},
\texttt{wipe\_desk}
} \\
\hline
\makecell{Correct grasp pose\\may not be found} & 
\makecell{
\texttt{change\_channel},
\texttt{change\_clock},
\texttt{hang\_frame\_on\_hanger},
\texttt{put\_knife\_in\_knife\_block},\\
\texttt{put\_knife\_on\_chopping\_board},
\texttt{screw\_nail},
\texttt{slide\_cabinet\_open},
\texttt{slide\_cabinet\_open\_and\_place\_cups},\\
\texttt{stack\_wine},
\texttt{take\_cup\_out\_from\_cabinet},
\texttt{turn\_tap},
\texttt{tv\_on}
}\\
\hline
\makecell{Hard for LLMs \\due to other reasons} 
& \makecell{
\texttt{beat\_the\_buzz},
\texttt{scoop\_with\_spatula}
}\\
\hline
\makecell{Brittle to \\explorations in RL} &
\makecell{
\texttt{hit\_ball\_with\_queue},
\texttt{hockey},
\texttt{insert\_usb\_in\_computer},
\texttt{move\_hanger},
\texttt{place\_cups},\\
\texttt{place\_hanger\_on\_rack},
\texttt{plug\_charger\_in\_power\_supply},
\texttt{pour\_from\_cup\_to\_cup},\\
\texttt{place\_all\_groceries\_in\_cupboard},
\texttt{put\_books\_on\_bookshelf},
\texttt{put\_bottle\_in\_fridge},\\
\texttt{put\_groceries\_in\_cupboard},
\texttt{put\_money\_in\_safe},
\texttt{put\_place\_in\_colored\_dish\_rack},\\
\texttt{put\_toilet\_roll\_on\_stand},
\texttt{put\_umbrella\_in\_umbrella\_stand},
\texttt{reach\_and\_drag},
\texttt{remove\_cups},\\
\texttt{set\_the\_table},
\texttt{setup\_checkers},
\texttt{setup\_chess},
\texttt{stack\_chairs},
\texttt{stack\_cups},
\texttt{sweep\_to\_dustpan},\\
\texttt{take\_frame\_off\_hanger},
\texttt{take\_money\_out\_safe},
\texttt{take\_plate\_off\_colored\_dish\_rack},
\texttt{water\_plants},\\
\texttt{weighing\_scales}
}\\
\hline
\end{tabular}
\end{table}

\begin{figure*}[!t]
\centering
\scalebox{1.0}{
\begin{subfigure}[b]{0.15\textwidth}
\centering
\includegraphics[width=\textwidth]{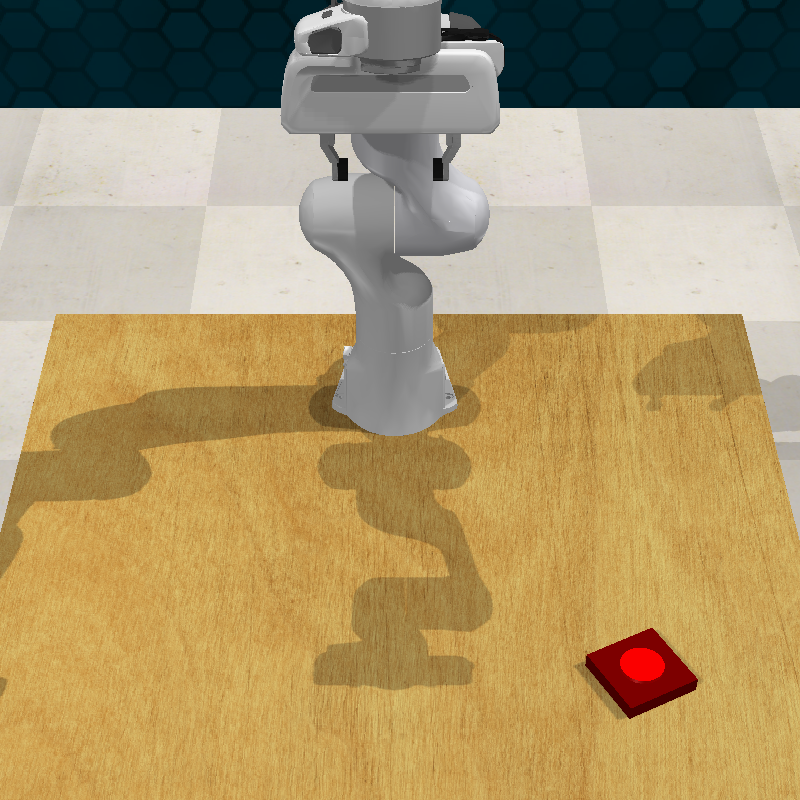}
{(a) \texttt{PB}}    
\end{subfigure}
\hspace{-0.1cm}
\begin{subfigure}[b]{0.15\textwidth}  
\centering 
\includegraphics[width=\textwidth]{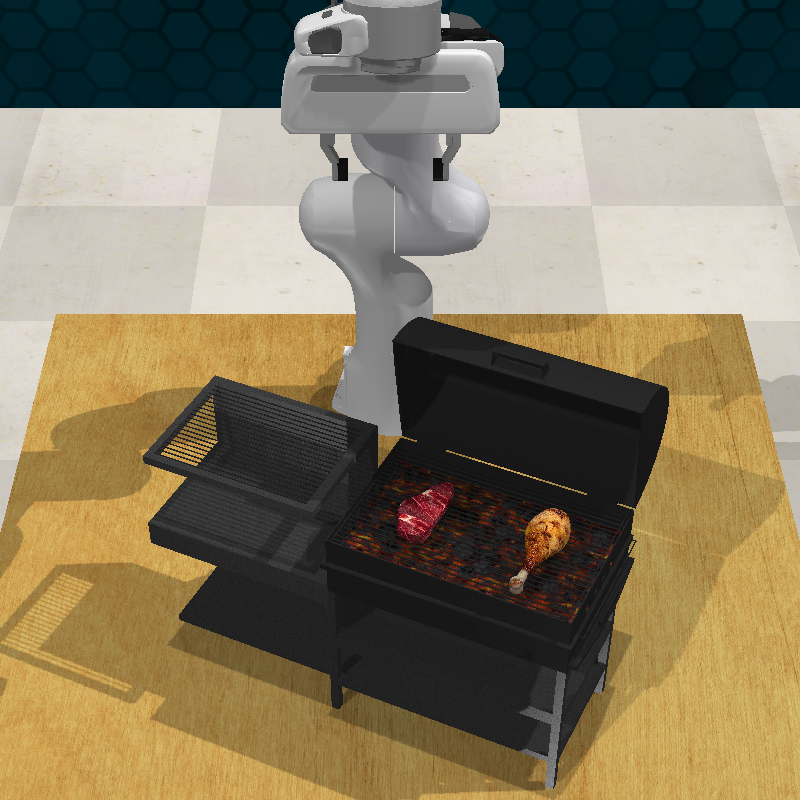}
{(b) \texttt{MOG}}      
\end{subfigure}
\hspace{-0.1cm}
\begin{subfigure}[b]{0.15\textwidth}   
\centering 
\includegraphics[width=\textwidth]{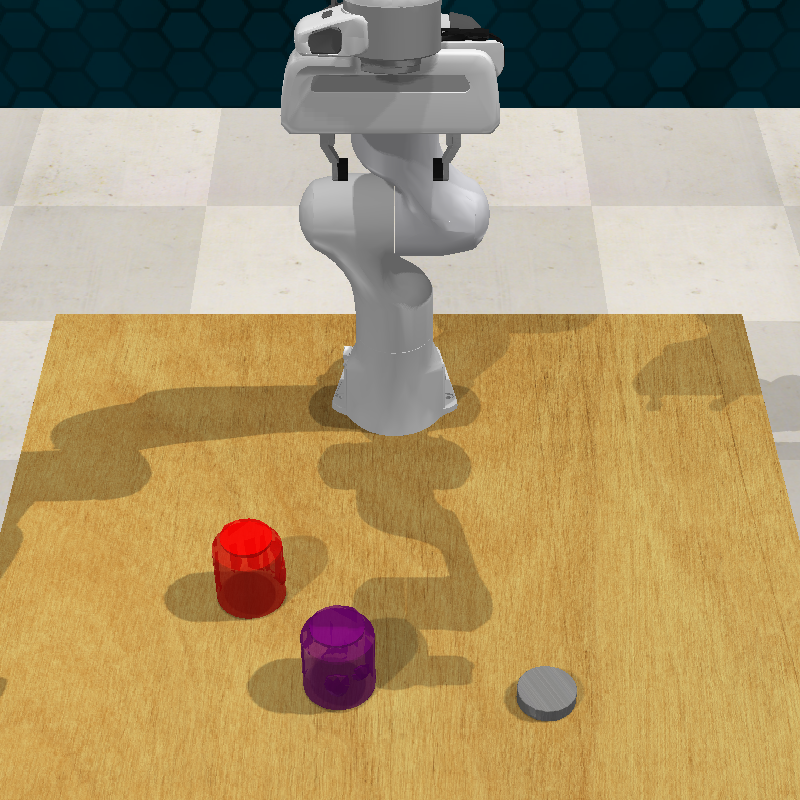}
{(c) \texttt{CJ}}    
\end{subfigure}
\hspace{-0.1cm}
\begin{subfigure}[b]{0.15\textwidth}   
\centering 
\includegraphics[width=\textwidth]{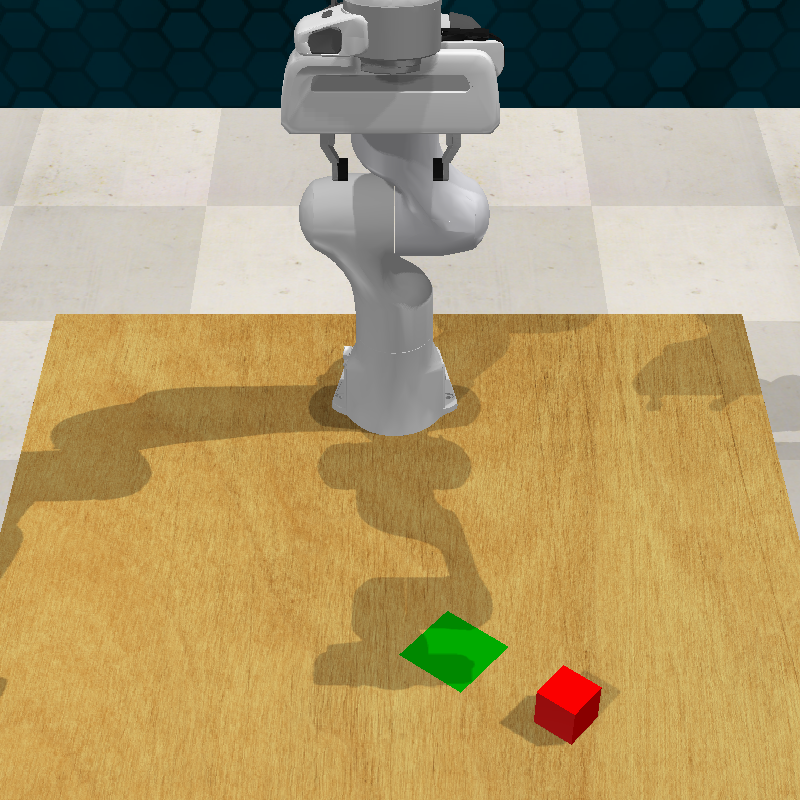}
{(d) \texttt{SBTT}}  
\end{subfigure}
\hspace{-0.1cm}
\begin{subfigure}[b]{0.15\textwidth}   
\centering 
\includegraphics[width=\textwidth]{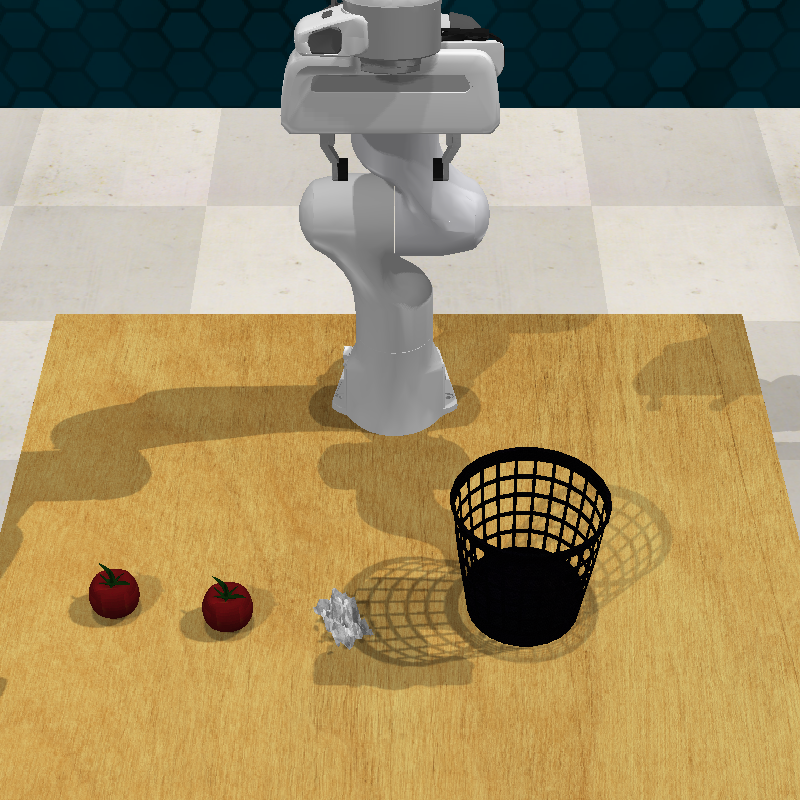}
{(e) \texttt{PICC}}     
\end{subfigure}
\hspace{-0.1cm}
\begin{subfigure}[b]{0.15\textwidth}   
\centering 
\includegraphics[width=\textwidth]{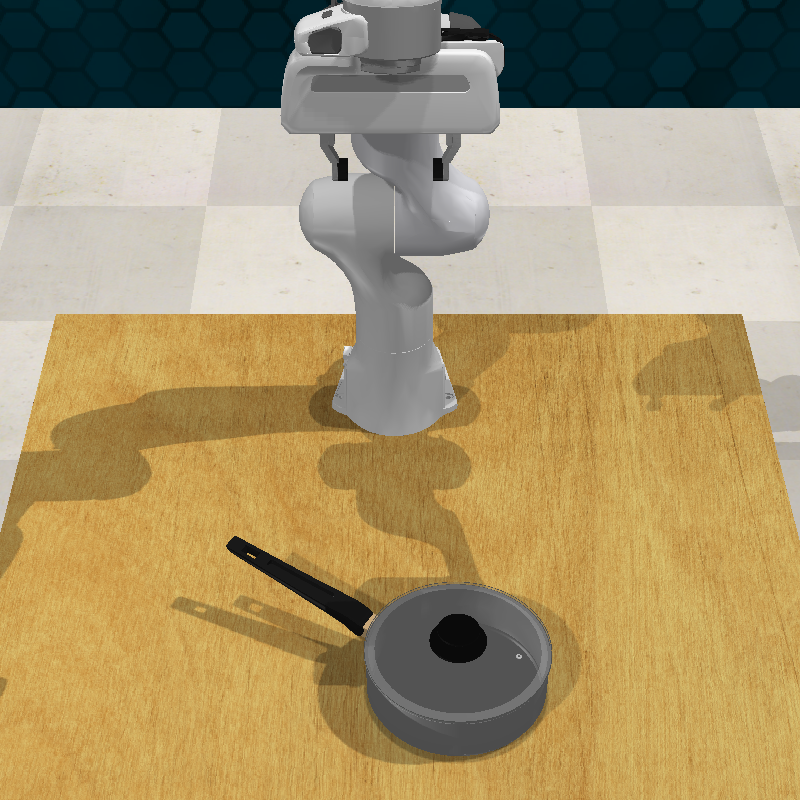}
{(f) \texttt{TLOS}}  
\end{subfigure}}
\vskip\baselineskip
\vspace{-0.3cm}
\scalebox{1.0}{
\begin{subfigure}[b]{0.15\textwidth}   
\centering 
\includegraphics[width=\textwidth]{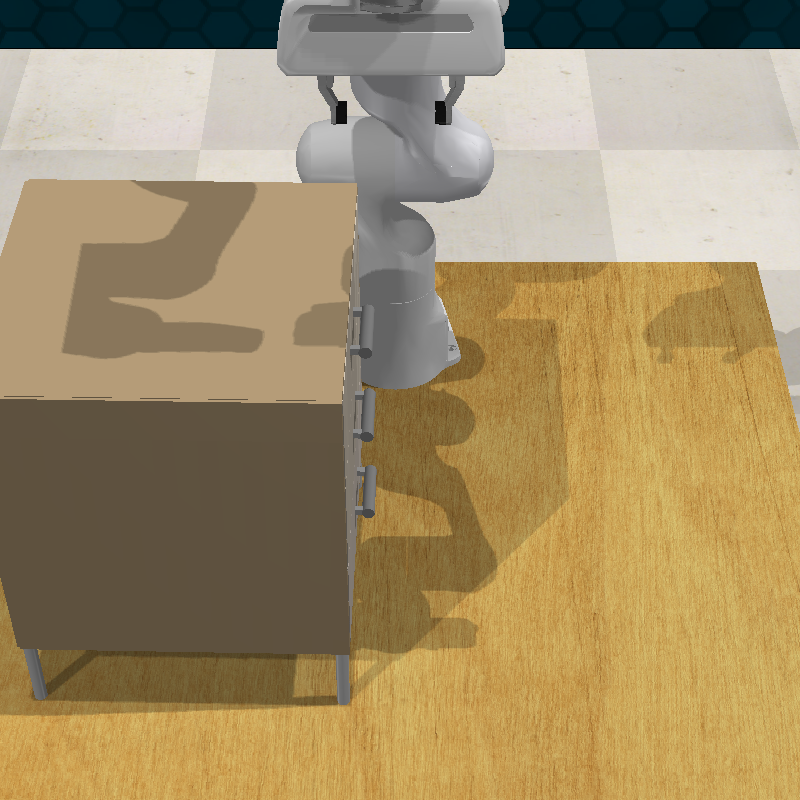}
{\color{black}(g) \texttt{OD}}  
\end{subfigure}
\hspace{-0.1cm}
\begin{subfigure}[b]{0.15\textwidth}   
\centering 
\includegraphics[width=\textwidth]{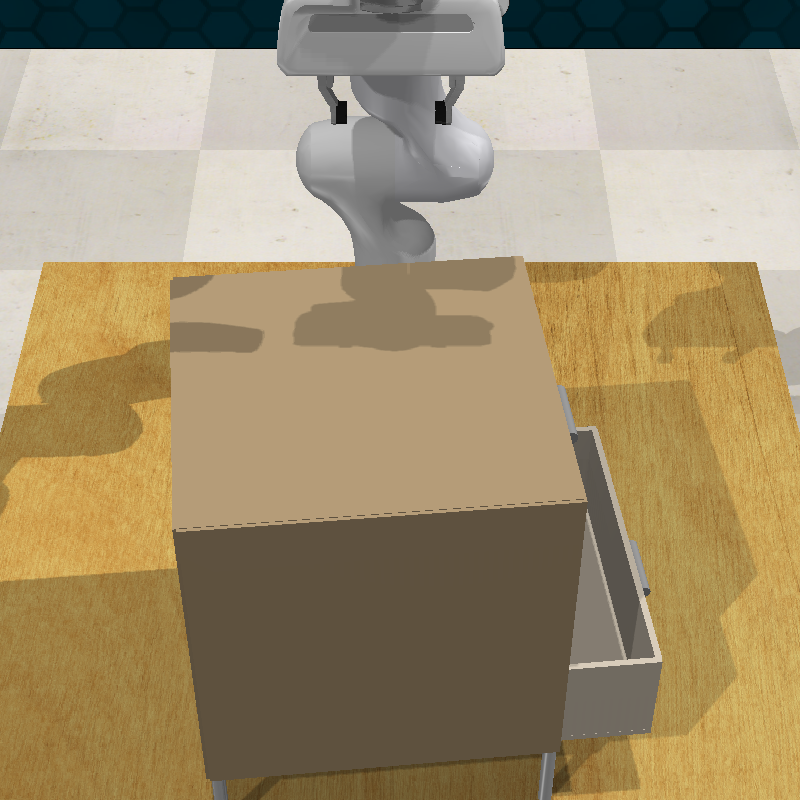}
{\color{black}(h) \texttt{CD}}     
\end{subfigure}
\hspace{-0.1cm}
\begin{subfigure}[b]{0.15\textwidth}   
\centering 
\includegraphics[width=\textwidth]{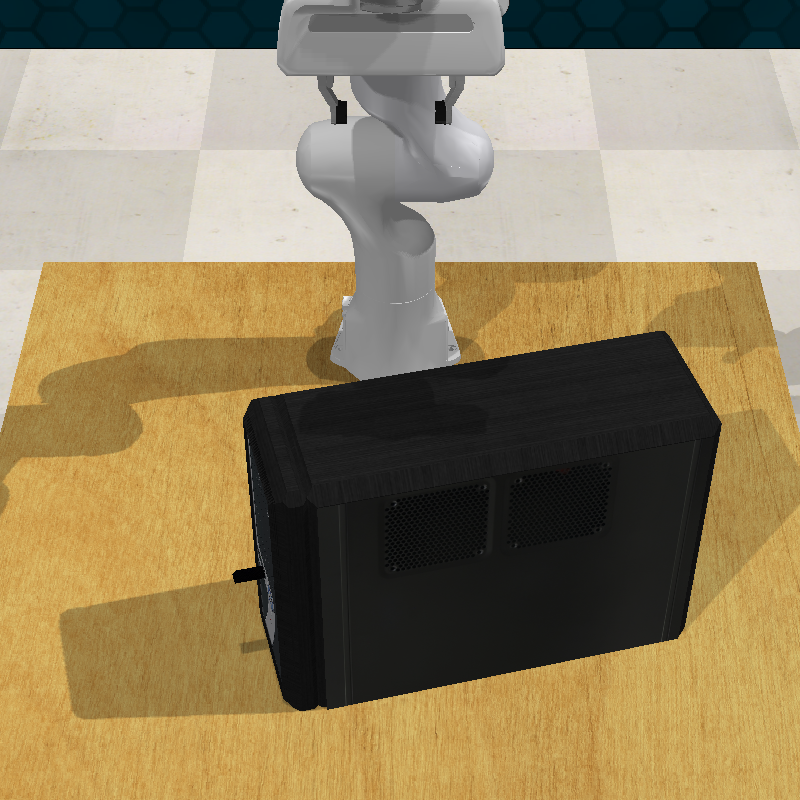}
{\color{black}(i) \texttt{TUOOC}}
\end{subfigure}
\hspace{-0.1cm}
\begin{subfigure}[b]{0.15\textwidth}   
\centering 
\includegraphics[width=\textwidth]{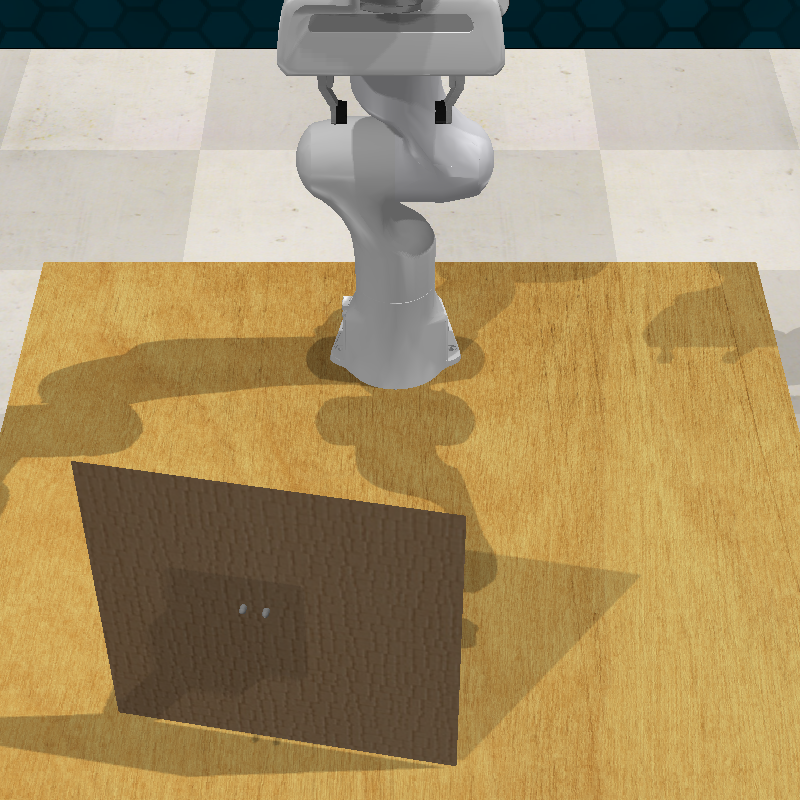}
{\color{black}(j) \texttt{UC}}  
\end{subfigure}
\hspace{-0.1cm}
\begin{subfigure}[b]{0.15\textwidth}   
\centering 
\includegraphics[width=\textwidth]{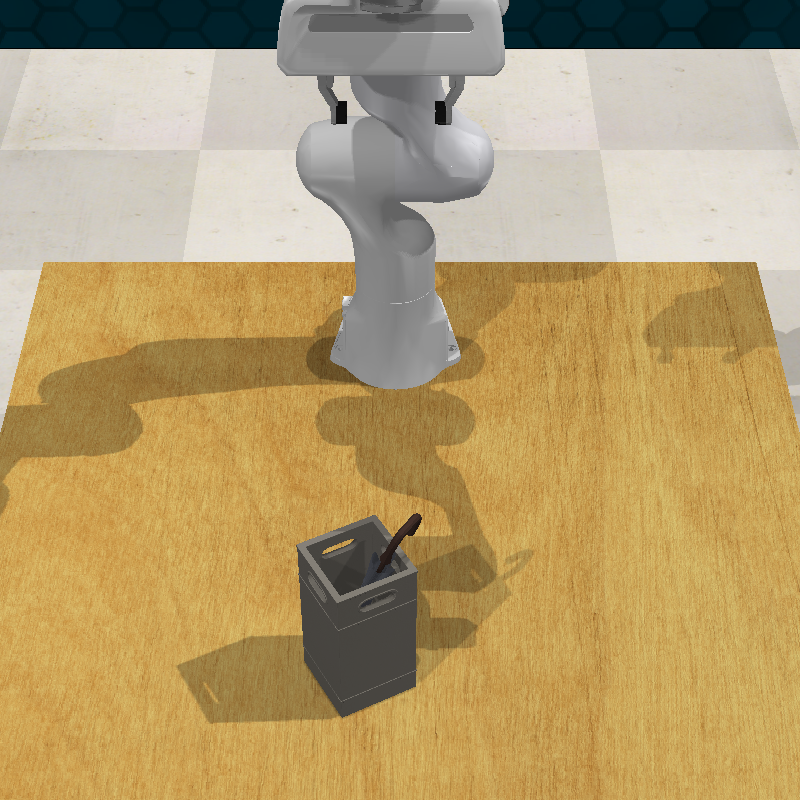}
{\color{black}(k) \texttt{TUOOUS}}
\end{subfigure}
\hspace{-0.1cm}
\begin{subfigure}[b]{0.15\textwidth}
\centering 
\includegraphics[width=\textwidth]{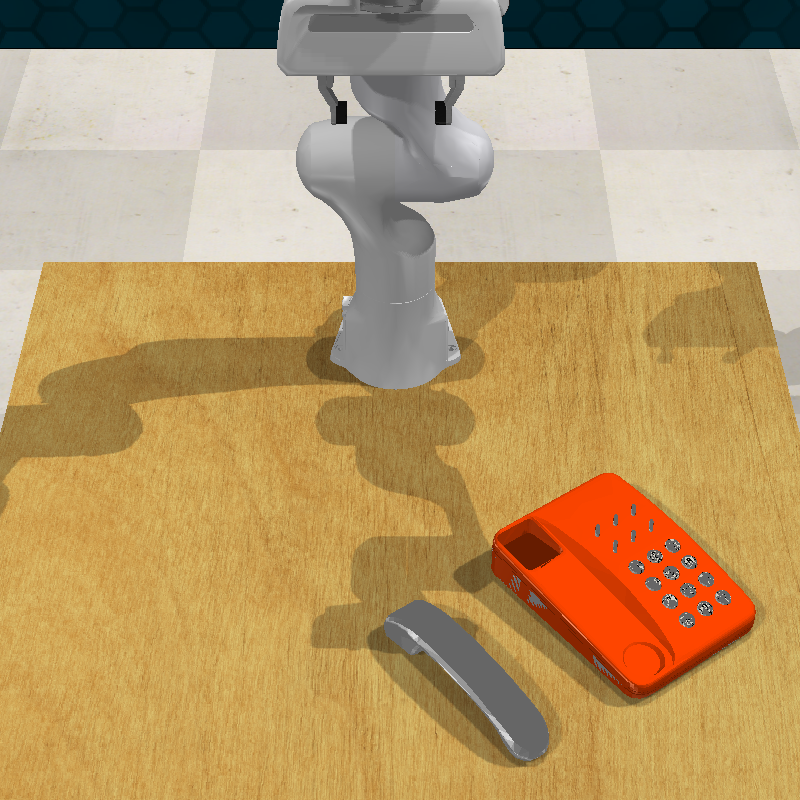}
{\color{black}(l) \texttt{POB}}  
\end{subfigure}}
\figuretag{A-1}
\caption{\color{black}The tasks in the {\texttt{RLBench}} environment: (a) \texttt{push\_button} (\texttt{PB});  (b) \texttt{meat\_off\_grill} (\texttt{MOG}); (c) \texttt{close\_jar} \\(\texttt{CJ}); (d) \texttt{slide\_block\_to\_target} (\texttt{SBTT}); (e) \texttt{put\_item\_in\_container} (\texttt{PIIC}); (f) \texttt{take\_lid\_off\_sauce-\\pan} (\texttt{TLOS}); (g) \texttt{open\_drawer} (\texttt{OD});  (h) \texttt{close\_drawer} (\texttt{CD}); (i) \texttt{take\_usb\_out\_of\_computer} (\texttt{TUOOC}); (j) \texttt{unplug\_charger} (\texttt{UC}); (k) \texttt{take\_umbrella\_out\_of\_umbrella\_stand} (\texttt{TUOOUS}); (l) \texttt{phone\_on\_base} (\texttt{POB})} 
\label{Fig2a}
\end{figure*}

These 12 \texttt{RLBench} tasks, visualized in Fig. \ref{Fig2a}, are introduced as follows. 

\begin{itemize}
\color{black}
\item [(a)] \texttt{push\_button}: The robot needs to push a button with its gripper.
\item [(b)] \texttt{meat\_off\_grill}: The robot needs to move a chicken off the grill.
\item [(c)] \texttt{close\_jar}: The robot needs to close a jar with a lid.
\item [(d)] \texttt{slide\_block\_to\_target}: The robot needs to slide a block to a target position. Compared with \texttt{PandaSlide} in \texttt{panda\_gym}, the friction between the block and table in the \texttt{RLBench} environment is larger so that the robot gripper may need to push the block multiple times in order to complete the task successfully.  
\item [(e)] \texttt{put\_item\_in\_container}: The robot needs to put an item in a container with its gripper.
\item [(f)] \texttt{take\_lid\_off\_saucepan}: The robot needs to take a lid off a saucepan.
\color{black}
\item [(g)] \texttt{open\_drawer}: The robot needs to open a drawer from the handle with its gripper.
\item [(h)] \texttt{close\_drawer}: The robot needs to close a drawer with its gripper.
\item [(i)] \texttt{take\_usb\_out\_of\_computer}: The robot needs to unplug a USB from a computer.
\item [(j)] \texttt{unplug\_charger}: The robot needs to unplug a charger and place it in a certain location with its gripper.
\item [(k)] \texttt{take\_umbrella\_out\_of\_umbrella\_stand}: The robot needs to take an umbrella out of an umbrella stand by lifting the umbrella.
\item [(l)] \texttt{phone\_on\_base}: The robot needs to pick a phone and place it on a phone base.
\end{itemize}

\subsubsection{\texttt{panda\_gym} Task Descriptions}
Five tasks in {\texttt{panda\_gym}} \cite{gallouedec2021panda} are considered in the simulation experiments of this paper. 

\begin{figure}[!t]
\centering
\scalebox{1.0}{
\begin{subfigure}[b]{0.18\textwidth}
\centering
\includegraphics[width=\textwidth]{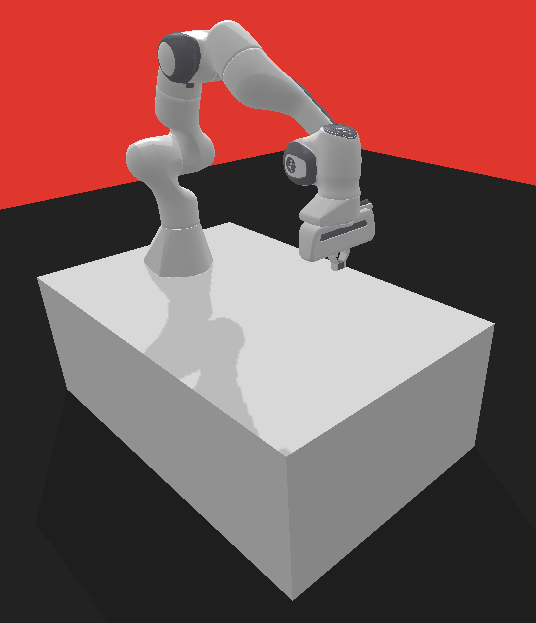}
{(a)}
\end{subfigure}
\hspace{-0.15cm}
\begin{subfigure}[b]{0.18\textwidth}
\centering 
\includegraphics[width=\textwidth]{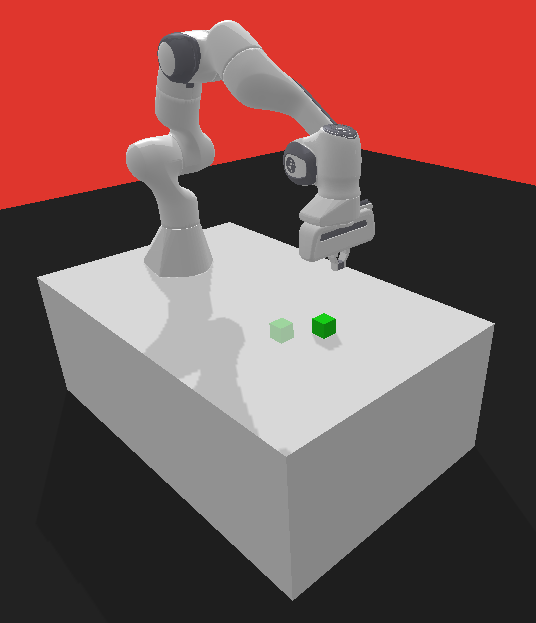}
{(b)}   
\end{subfigure}
\hspace{-0.15cm}
\begin{subfigure}[b]{0.18\textwidth}   
\centering
\includegraphics[width=\textwidth]{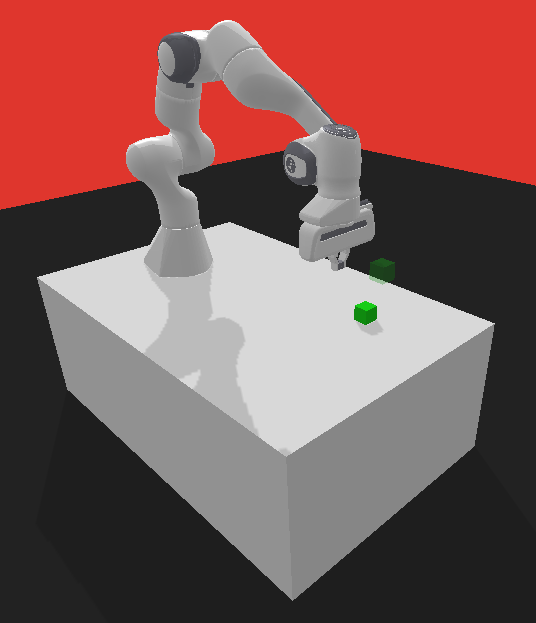}
{(c)} 
\end{subfigure}
\hspace{-0.15cm}
\begin{subfigure}[b]{0.18\textwidth}   
\centering 
\includegraphics[width=\textwidth]{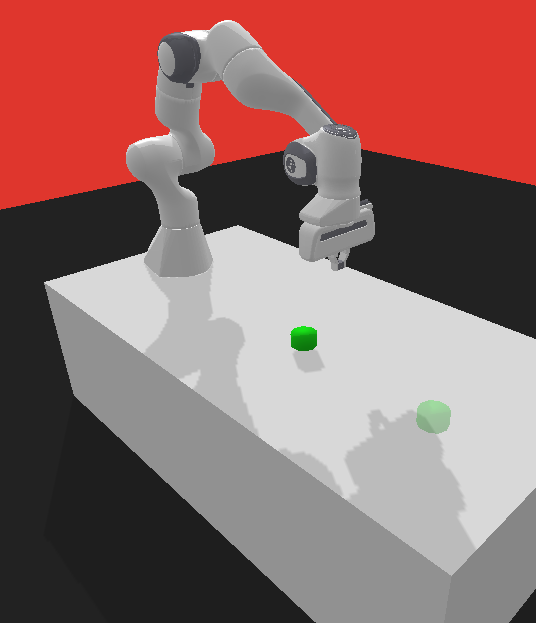}
{(d)}  
\end{subfigure}
\hspace{-0.15cm}
\begin{subfigure}[b]{0.18\textwidth}   
\centering 
\includegraphics[width=\textwidth]{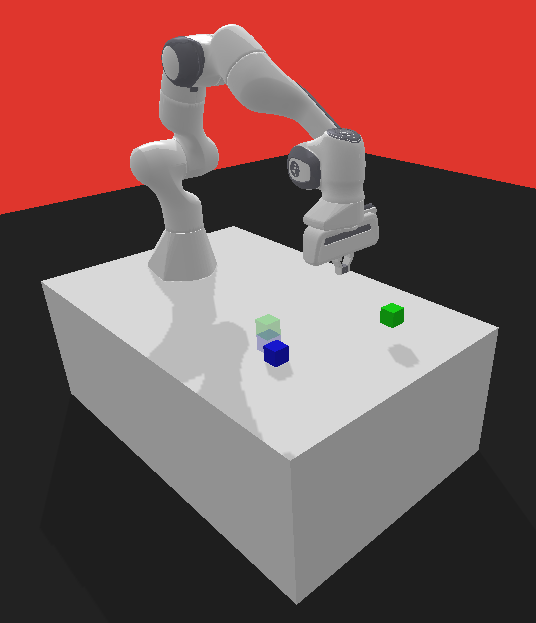}
{(e)}  
\end{subfigure}}
\figuretag{A-2}
\caption{\color{black}The five tasks in the {\texttt{panda\_gym}} environment: (a) \texttt{PandaReach}; (b) \texttt{PandaPush}; (c) \texttt{PandaPickAndPlace}; (d) \texttt{PandaSlide}; (e) \texttt{PandaStack} (dark color: cube(s); shaded color: the target position(s))} 
\label{Fig1a}
\end{figure}
\color{black}
\begin{itemize}
\item [(a)] \texttt{PandaReach}: The robot gripper needs to reach a point in the space.
\item [(b)] \texttt{PandaPush}: The gripper needs to push a cube from one location to another location {\textit{within}} the reachable region of the gripper.
\item [(c)] \texttt{PandaPickAndPlace}: The robot needs to pick a cube from one location and place it to another with its gripper.
\item [(d)] \texttt{PandaSlide}: The gripper needs to slide a cube from one location to another location {\textit{beyond}} the reachable region of the gripper.
\color{black}
\item [(e)] \texttt{PandaStack}: The gripper needs to stack two cubes by picking and placing them to assigned locations.
\end{itemize}

\section{Experiment Hyperparameters}
\label{SB}
TABLE \ref{TA1} shows the hyperparameters of the RLingua with TD3 for the tasks in both \texttt{panda\_gym} and \texttt{RLBench} environments. The meanings of the notations are given in Algorithm \ref{Alg1}. We use the same values of $\gamma$ for the standard TD3 algorithm in consideration of fair comparisons. The other hyperparameters not listed in TABLE \ref{TA1} are the same as those in \texttt{stable\_baselines3} (See \url{https://github.com/DLR-RM/stable-baselines3}). 

\begin{table}[!h]
\centering
\captionsetup{justification=centering}
\tabletag{A-II}
\caption{The experiment hyperparameters}
\label{TA1}
\begin{tabular}{ccwc{1cm}wc{1cm}wc{1cm}wc{1cm}}
\hline
Environment & $\gamma$ & $p^{\mathrm{LLM}}$ & $\lambda_{\mathrm{annl}}$ & $\lambda_{\mathrm{IM}}$\\
\hline
\texttt{panda\_gym} & 0.975 & 0.25& 0.99995 & 1.0\\ 
\texttt{RLBench} & 0.96 & 0.25& 0.999999 & 1.0 \\  
\hline
\end{tabular}
\end{table}

\section{Discussions}
\label{SD}
This paper introduces the RLingua framework, which utilizes large language models (LLMs) to assist in reinforcement learning (RL). The main storyline of this paper is that the extracted rule-based controllers from LLMs are helpful in reducing the sample complexity of RL. We take the initial step in this direction, demonstrating the promising aspects of RLingua in various robot tasks. With the rapid development of LLMs, we believe that future LLMs can be prompted to generate reasonable rule-based controllers for more complicated robot tasks, for which the framework of RLingua can also be used to improve the sample efficiency of RL. On the other storyline that we did not emphasize in this paper, we have to admit that there are always some complicated robot tasks that cannot be solved perfectly by only prompting LLMs, even if we hold an optimistic attitude about the rapid development of LLMs. In this perspective, RLingua provides a method to improve the imperfect LLM-generated controllers by RL. We hope that our work in this paper can motivate more relevant research in both storylines. 

In the scope of this paper, we assume that the Franka Emika Panda robot can only implement some basic movements, such as displacements and rotations in a small region relative to the current gripper position and orientation. Note that the learned robot primitives in this work can be further extended to more high-level robot tasks \cite{huang2022language, ahn2022can, huang2022inner, zhang2023bootstrap, wake2023chatgpt}. In addition, the framework of RLingua can be combined with the existing methods that use LLMs to assist the reward design \cite{kwon2023reward, yu2023language, xie2023text2reward, ma2023eureka}. Since both reward design and data efficiency are key practical challenges of RL, this work steps forward to obtain more practical RL agents by leveraging LLMs. 

Furthermore, the LLM prompt design described in Section \ref{S31} aims to incorporate the prompt sub-parts requiring expert knowledge to a fixed template and leave the generated task descriptions and human feedback, which can be finished by non-experts, to the users. In this sense, we take a step further to achieve a personalized robot controller that can be obtained without expert knowledge. 

{\color{black}We would like to note that the framework of RLingua in this paper needs an external perception to extract the state information of robot tasks. The future research direction, stimulated by the development and increased affordability of multi-modal LLMs \cite{yang2023dawn}, involves developing an end-to-end RLingua-like framework.}

\end{document}